\documentclass[twocolumn]{article}
\usepackage[round]{natbib}
\usepackage{graphicx}
\usepackage{amsmath}

\usepackage[margin=0.6in]{geometry}

\usepackage{moreverb,url}
\usepackage[T1]{fontenc}
\usepackage{Alegreya} %% Option 'black' gives heavier bold face 

\usepackage[colorlinks,bookmarksopen,bookmarksnumbered,citecolor=red,urlcolor=red]{hyperref}

\usepackage[usenames,dvipsnames,svgnames,table]{xcolor}

\usepackage{algorithm}
\usepackage[noend]{algpseudocode}
\floatname{algorithm}{Procedure}
\usepackage{algorithmicx}
\newcommand\algcomment[1]{ \textcolor{gray}{//#1}}

\newcommand\scalemath[2]{\scalebox{#1}{\mbox{\ensuremath{\displaystyle #2}}}}

\newcommand\BibTeX{{\rmfamily B\kern-.05em \textsc{i\kern-.025em b}\kern-.08em
T\kern-.1667em\lower.7ex\hbox{E}\kern-.125emX}}

\newcommand\blfootnote[1]{%
  \begingroup
  \renewcommand\thefootnote{}\footnote{#1}%
  \addtocounter{footnote}{-1}%
  \endgroup
}

\begin{document}

\title{On the mechanical contribution of head stabilization to passive dynamics of anthropometric walkers}

\author{Mehdi Benallegue$^{1,2}$, Jean-Paul Laumond$^3$ and Alain Berthoz$^4$}

\maketitle

\blfootnote{$^1$  CNRS-AIST JRL (Joint Robotics Laboratory), IRL}

\blfootnote{2 National Institute of Advanced Industrial Science and Technology (AIST)}

\blfootnote{$^3$ Joint research unit CNRS/ENS/INRIA/PSL UMR8548, Paris. (this work has been done when the author was with LAAS-CNRS in Toulouse).}

\blfootnote{$^4$ Coll\`ege de France, Paris.}

\blfootnote{\\This work is supported by
\begin{itemize}
\item Grants from the EU Project CLONS (Closed-Loop Neural Prosthesis for Vestibular Disorders, Grant Agreement 225929)
\item The French Romeo2 project.
\item The European Research Council grant Actanthrope (ERC-ADG 340050).
\item The French government under management of Agence Nationale de la Recherche as part of the ``Investissements d'avenir''   program, reference ANR19-P3IA-0001 (PRAIRIE 3IA Institute).
\end{itemize} }

\begin{footnotesize}
\begin{abstract}
{\small During the steady gait, humans stabilize their head around the vertical orientation. While there are sensori-cognitive explanations for this phenomenon, its mechanical effect on the body dynamics remains unexplored. In this study, we take profit from the similarities that human steady gait share with the locomotion of passive dynamics robots. We introduce a simplified anthropometric 2D model to reproduce a broad walking dynamics. In a previous study, we showed heuristically that the presence of a stabilized head-neck system significantly influences the dynamics of walking. This paper gives new insights that lead to understanding this mechanical effect. In particular, we introduce an original cart upper-body model that allows to better understand the mechanical interest of head stabilization when walking, and we study how this effect is sensitive to the choice of control parameters.}

\end{abstract}
{Humanoid Robots, Legged Robots, Biomimetics}
\end{footnotesize}
%\keywords{Head stabilization, Human locomotion, humanoid robotics, biomechanics, passive-dynamic walkers}

\section{Introduction}
Bipedal locomotion covers various kinds of walking behaviors. For robots, it ranges from quasi-static locomotor sequences through dynamical fully actuated walking pattern generations to the motion of passive walking mechanical systems also called passive-dynamic walkers~\citep{wieber2015modeling}. While the first ones require a big amount of energy to ensure locomotion, the latter need no external energy on shallow slopes, but without a heavyweight trunk~\citep{Collins2005science}, and if equipped with weak actuators, they can have an upright trunk and face flat grounds with human-like energy efficiency~\citep{Alexander2005science}. 

Humans also have different walking strategies depending on their environment, mostly the ground roughness, that we call also \textit{texture} of the ground~\citep{jpl2017ijrr}. We define a textured ground by a surface for which the unevenness follows a probability distribution. On high textures, such as uneven terrain, walking requires anticipation for the stepping position, which uses cognitive resources and causes cautious, highly stiff and energy-consuming motions. When walking on flat grounds humans don't need attention on their foot placement, walking control is then mainly generated within the spinal cord rather than in the brain~\citep{Dietz2003cn}. This strategy is usually modeled by the combination of a rhythm generator called Central Pattern Generator and reflexes in response to peripheral stimuli. We call this control ``steady gait''. Steady-gait-based walking is more energy-efficient and takes maximum advantage of the natural passivity of the mechanical structure of the body. The dependency on the passive dynamics and the weakness of actuation make this walking control sensitive to the environment. If the texture is too uneven, humans switch to the robust predictive mode. Similarly, passive dynamics robots are not stable on too highly textured grounds and eventually fall. Passive dynamics robot walking and human steady gait share some other characteristics, such as their periodic dynamics and the natural attractive limit-cycle~\citep{Goswami1997ar}. Passive dynamic robots are also believed to produce visually more human-like motions~\citep{Collins2005science}. 

%For this kind of gaits, robotics traditional balance metrics such as the zero moment point~\cite{vukobratovic2004ijhr} are not relevant. For this reason

These gaits emerge from the dynamics of all body parts. Each limb or joint influences the motion according to its inertia and applied forces, either subject to its passivity or under a specific actuation. Therefore, several researchers used the resemblance between walking humans and passive dynamic robots for studying the dynamical contribution of several features of human gait~\citep{kaddar2015ras,Wisse2006humanoids,Chevallereau2009itro}. These studies aim at understanding the effect of these features and their importance in human gait and enable to exploit them to improve the performances of the robotic walkers.

Among the recognized features of human gait lies one important phenomenon: active head tilt stabilization. Head stabilization is believed to offer a stable and consistent egocentric reference frame for motion perception and control~\citep{Berthoz2000book}. %(See Figure~\ref{fig:HeadStabilization}). 
A more recent result shows also that head stabilization improves the estimation of the vertical direction by allowing to resolve the gravito inertial ambiguity~\citep{farkhatdinov2019gravito}. 

This stabilization is known to be particularly important for locomotion~\citep{Pozzo1990ebr}. The stability of the head increases the gaze efficiency, for example through vestibulo-ocular reflex, into anticipating the future paths~\citep{Hicheur2005nl}. This allows preserving the energy-efficient gait on various kinds of real-world environments~\citep{matthis2018gaze}.
The effort for stabilization is known to be increased particularly when balance becomes more challenging such as for elderly people~\citep{cromwell2002influence}, or when subject to a disturbance~\citep{Cromwell2001SagittalPH,cromwell2003movement,cromwell2004head}, even when the spinal segments were stiffened into a "rigid" body, requiring compensation from the neck and the pelvis joints~\citep{nadeau2003head}.

%\begin{figure}
%\begin{center}
%\includegraphics[width=\columnwidth]{img/HeadStabilization}
%\caption{\label{fig:HeadStabilization} Human locomotion and head stabilization. In A stick figures show
%a captured walking motion. In B all the figures were combined by translation
%so that the auditory meatus superpose. Notice the small variation of the head
%angular position. In C is an enlarged version of the B figure. Courtesy of
%Amblard et al~\cite{Amblard1988inbook}}
% \end{center}
% \end{figure}

In this paper, we aim at understanding what is the mechanical effect of the control of the head on the dynamics of both the upper body and the whole body gait. Indeed, the head represents 8\% of total body mass and occupies the top 12\% of the body~\citep{Armstrong1988report}. The head accounts then for an important contribution to the angular momentum of the body relative to the stance foot position. The dynamics of head stabilization should then have noticeable effects on gait motion\footnote{The angular momentum of a body around the contact point is the cross product of the velocity vector 2and the distance to the contact point, all multiplied with the mass. The head, being at the longest distance to the contact, and given its mass, tells us about the importance of its contribution to the total angular momentum.}.

However, to our best knowledge, no work attempted to understand the effects of the control of the head-neck system on the gait dynamics. In fact, classic studies of the dynamics of gait neglect this effect~\citep{delp1990interactive,anderson2001dynamic,thelen2006using,skalshoi2015walking}.

In a former work~\citep{jpl2017ijrr}, we have shown that a stabilized upper-body in an anthropometric model contributes to stabilizing walking dynamics compared to a walker with a rigid neck.

This paper aims at studying more in-depth the mechanical effect of head stabilization on the dynamics of gait. We first introduce two walker models. Both walkers share the same mass distribution. In the first model the torso, the neck, and the head constitute a single rigid body. In the second model the torso, the neck, and the head constitute a two-degree-of-freedom articulated chain. Then we recap the heuristic approach introduced in~\citep{benallegue2013rss}. It is based on numerical simulation and it shows that the average number of successful steps on uneven grounds significantly increases when the head is stabilized. 

The first contribution of this paper is to give an explanation of the head stabilization effect in terms of energy consumption during the steady gait and its effect on the swing leg dynamics. 

The second contribution is the study of an original cart upper-body model that allows accounting for the impacts within the limit-cycle. The head-neck system appears to play the role of a low-pass filter on the force dynamics of the upper-body. 

The third contribution is the study of the sensitivity of this effect to control parameters. By generating random parameters, we show that head stabilization provides a significant contribution to the viability of a stable gait on flat ground. We show also that it improves the overall robustness of the control with regard to ground perturbations.

Afterward, the consequences and perspectives of these results are presented.

 The details of the actuation and impact models, as well as the technical developments of the cart upper body model linearization,  are reported in the supplementary material.

\section{Passive walker models, ground texture and simulations}\label{sec:models}
This section introduces the walker models we consider and gives an overview of the results presented in~\citep{jpl2017ijrr}.

\subsection{The kinematic and dynamic model}
\begin{figure}[]
\begin{center}
\includegraphics[width=0.5\columnwidth]{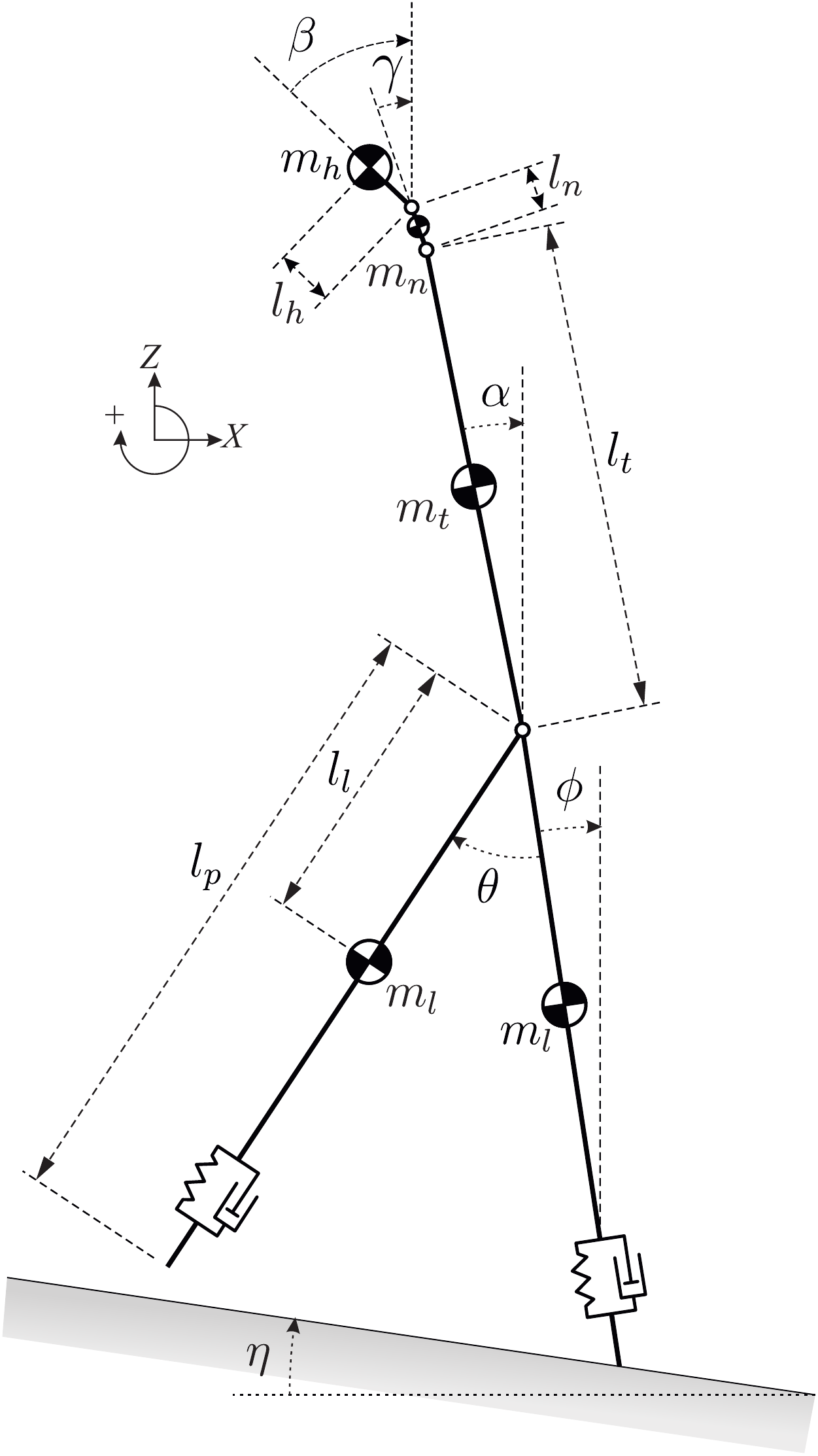}
\end{center}
\caption{ 
A representation of the model we simulate. The A model is the same structure subject to the constraints $\alpha=\beta=\gamma$. The B model has stabilized neck joints.
}
\label{Model}
\end{figure}

Our passive-dynamics walker model is a planar 5 limbs kinematic tree which operates in the sagittal plane. The limbs are: a head with a mass on the top;  a neck and a torso with masses at the middle; and the (knee-free) legs, each of which with a mass at distance $l_l$ from the hip. Every two successive limbs are attached with a rotational joint (see Fig~\ref{Model}). Each leg is ended with a prismatic joint equipped with a spring-damper. We call \emph{toe}, the bottom of the mobile part of the leg. The hip-toe length $l_p$ at the rest position of the spring is denoted by $l_{p,0}$. Note that the neck is modeled as an articulated body and not as a simple joint. This setting reflects the property of the head-neck system to have two centers of rotation in the sagittal plane: one at the base of the neck and the other at ear level~\citep{Viviani1975bc}.

Arms have an important impact when 3D dynamics is considered~\citep{Chevallereau2009itro}. However, since our model is in the sagittal plane, the influence of symmetric swinging arms should be minor and is then neglected. 

The total height of the robot is 1.91 m and the total mass is 80 Kg~\footnote{The actual simulation were performed with a model 10 times lighter to improve numerical conditioning, but all the dynamics is preserved since it is only a scale factor of the actual one}. The segment length and mass relative distribution are anthropometric~\citep{Armstrong1988report}.

It is important to emphasize that we do certainly \emph{not} aim at reproducing perfectly the human gait. Indeed, up to now, only simple dynamical models allow to recreate locomotion gaits \citep{Mombaur2009robotica} and accurate dynamical modeling of human walking is out of reach of current simulators. Here we model only a broad dynamics of an anthropomorphic walker in a steady limit-cycle gait, similarly to what is done in~\citep{donelan2002expbio,roos2011jbiomeca,tlalolini2011human}. Of course, any claim made on human dynamics requires a dedicated experimental validation on human subjects. However, modeling can be an important step to guide the research on humans by highlighting phenomena which would be more difficult to identify for humans.
 % These additions are active research topics, but their effects and dynamics are beyond the scope of this paper. We refer then to~\cite{McGeer1990ijrr, Collins2005science, van1999passive, su2007dynamic,McGeer1988report,benallegue2013icra} for dedicated studies.

\subsection{Two actuation models}
 In order to see the effect of head stabilization, we simulate the presented walker with two different actuation models. The first walker (Model A) we consider, has a rigid neck, i.e. the torso, the neck, and the head make a single rigid body. The second walker corresponds to the model of head stabilization (Model B): the neck joints are controlled to maintain a zero tilt for the head. 
 
 Apart from the neck, both walkers have the same controls: the torso is actuated to be stabilized upright while a lightweight controller actuates the inter-leg angle. Finally, a velocity driven foot impulsion is given just before the swing phase. Except for toe-off impulsion, the controllers for the robot are proportional-derivative (PD), each of them has two gain parameters. The gains for the lower body are chosen to be lightweight, to approach the low energy consumption of human's steady walk. The weakness of the control of the lower limbs make them sensitive to perturbations, and their dynamics can differ according to upper-body control. The upper body has to guarantee a successful vertical stabilization for the trunk and the head and has, therefore, stiffer actuators. 

%There are two ways for obtaining the values of the control gains. The first and the best one is to use optimization techniques to find out the best parameters of each model according to a balance criterion. However, our balance criterion, described in Subsection~\ref{subsec:mfpt}, is too long to compute and optimization would take years. The second method, to which we have to resort, is to take the values arbitrarily in the set of values guaranteeing stable steady gait on flat ground. If for all the samples, the same effect is qualitatively observed, we can safely generalize it to all the set. In our tests, among all the used values, all the described effect remain visible. 

Each PD controller has two gain parameters. The parameters we present here are the same that were used to obtain our previous results on uneven ground~\citep{jpl2017ijrr}. A detailed description of the gains is given in the supplementary material. Nevertheless, in the next section, we study how different values of these gains influence this dynamical effect.

\begin{table}
\begin{center}
\caption{
The parameters of our simulated walker}
\begin{tabular}{|c|c|c|}
\hline
& Length (m) & Mass (Kg)\\
\hline 
Head & $l_h$=0.09 & $m_h$ = 4 \\
Neck & $l_n$=0.07 & $m_n$ = 1 \\
Torso & $l_t$=0.75 &  $m_t$ = 45 \\
Leg & $l_{p,0}$=1&  $m_l$ = 15 \\
Leg CoM&$l_l$ = 0.40& \\ 
\hline 
\end{tabular}
\label{tbl:values}
\end{center}
\end{table} 

\subsection{Simulation and limit-cycle}
Our numerical simulations are performed in a tailored C++ framework for simulating passivity based walkers. The resolution of dynamics integration is achieved using the dynamic simulator Open Dynamics Engine (ODE). The simulation time-sample is of 1 ms, but a special simulation time-sample is set to 0.1 ms at impact instants in order to increase the physical realism. 

During the motion, we consider the state vector $\xi=( \theta, \dot{\theta},\dot{\phi},\eta ,\alpha ,\dot{\alpha},\gamma ,\dot{\gamma} ,\beta, \dot{\beta},l_p ,\dot{l_p})$ in the state space $\mathcal{S}$ evolving in time. The dynamics of walking systems is cyclic. At each step, there is an impact and a swing phase.

By simulating both models A and B we see that starting from the appropriate state their gaits are stable and balanced on flat ground. Furthermore, the motion is attracted to a stable limit-cycle. To obtain the values of these limit-cycles, we simulate the walking motion on a flat ground starting from 8000 uniformly distributed state values in the state-space. The simulation runs until it falls or it converges to a limit-cycle. If a simulated motion does not fall, we assume that an accurate convergence to the limit cycle happens after 400 steps. %In a perfect environment and simulation, the walker's state is supposed to converge to a unique limit value $\xi_l$. But due to numerical rounding and to simulation errors, the states have still varying values even after the convergence. So 
For each viable sample, we record the states between the 400th and the 500th step. This gives a set of $\tilde{\xi}_i$ of recorded states after the convergence. 
%This set of states occupies a small area in the state space within which the motion is no longer attracted to a limit-state $\xi_l$, but it generates a local chaotic behavior. We call this area the \emph{limit kernel}. This area is attractive, in the sense that, on flat ground, starting from the outside and in case the walker doesn't fall, the state will certainly enter the area in a finite number of steps. Moreover, as soon as the state is inside the limit kernel, it will never leave it in the absence of disturbances. 
In the absence of a perfectly accurate value for the limit cycle, we approximate $\xi_l$ by  the mean value of the recorded states$\{\tilde{\xi}_i\}$
(see Table~\ref{tbl:limitcycle}).

\begin{table}
\caption{
 The value of the limit-cycle state for each model.}
\begin{tabular}{|l|c|c|}
\hline
& Rigid neck (Model A) & Head stab. (Model B)\\
\hline 
$  \theta$ (rad) &$    6.23\times 10^{-1}$& $ 5.17\times 10^{-1}$\\
$\dot{\theta}$     (rad/s)&$    -2.64\times 10^{-1    }$&$ 1.07$\\
$\dot{\phi}$     (rad/s)&$    1.58 $&$ 1.32 $\\
$\eta$             (rad)&$    0    $&$ 0 $\\
$\alpha $        (rad)&$    -2.88\times 10^{-2    }$&$ -2.89\times 10^{-2}$\\ 
$\dot{\alpha}$    (rad/s)&$    -2.83 \times 10^{-1    }$&$ -2.77\times 10^{-1}$\\
$\gamma$         (rad)&-&$ -2.60\times 10^{-2}$\\
$\dot{\gamma}$    (rad/s)&-&$ -6.60\times 10^{-2}$\\
$\beta$            (rad)&    -&$ -4.49\times 10^{-3}$\\
$\dot{\beta}$    (rad/s)&-&$-1.32\times 10^{-2}$\\
$l_p $            (m)&$    -1.12 \times 10^{-2    }$&$-1.28\times 10^{-2}$\\
$\dot{l_p}$        (m/s)&$    1.53 \times 10^{-2    }$&$4.40\times 10^{-3} $\\
\hline 
\end{tabular}
\label{tbl:limitcycle}
\end{table}

We see in this table that the values are globally close between the models, particularly the inclination of the trunk segment, but with few notable exceptions. First, the head ($\beta$ for Model B and $\alpha$ for Model A) is more vertical for the head stabilization model, which is the purpose of this model. Second, the lower limb values are relatively different with 18\% longer steps for Model A, and even a different sign for the inter-leg angular velocity. This discrepancy is only due to the difference between the two models in the actuation of the head-neck system. We recall in the next section our previous result on the consequences of this difference on simulated uneven ground, and we investigate afterward this effect more deeply.

\subsection{A model for ground texture}
Both models have stable limit-cycle gaits on flat ground. This gives the walkers equivalently perfect balance performances on this ideal environment. However, the dynamics of Model A and Model B differ because of head stabilization. The limit cycle is different and more importantly, the robustness of the walkers to a textured ground could also be different.

\begin{figure}
\begin{center}
\includegraphics[width=\columnwidth]{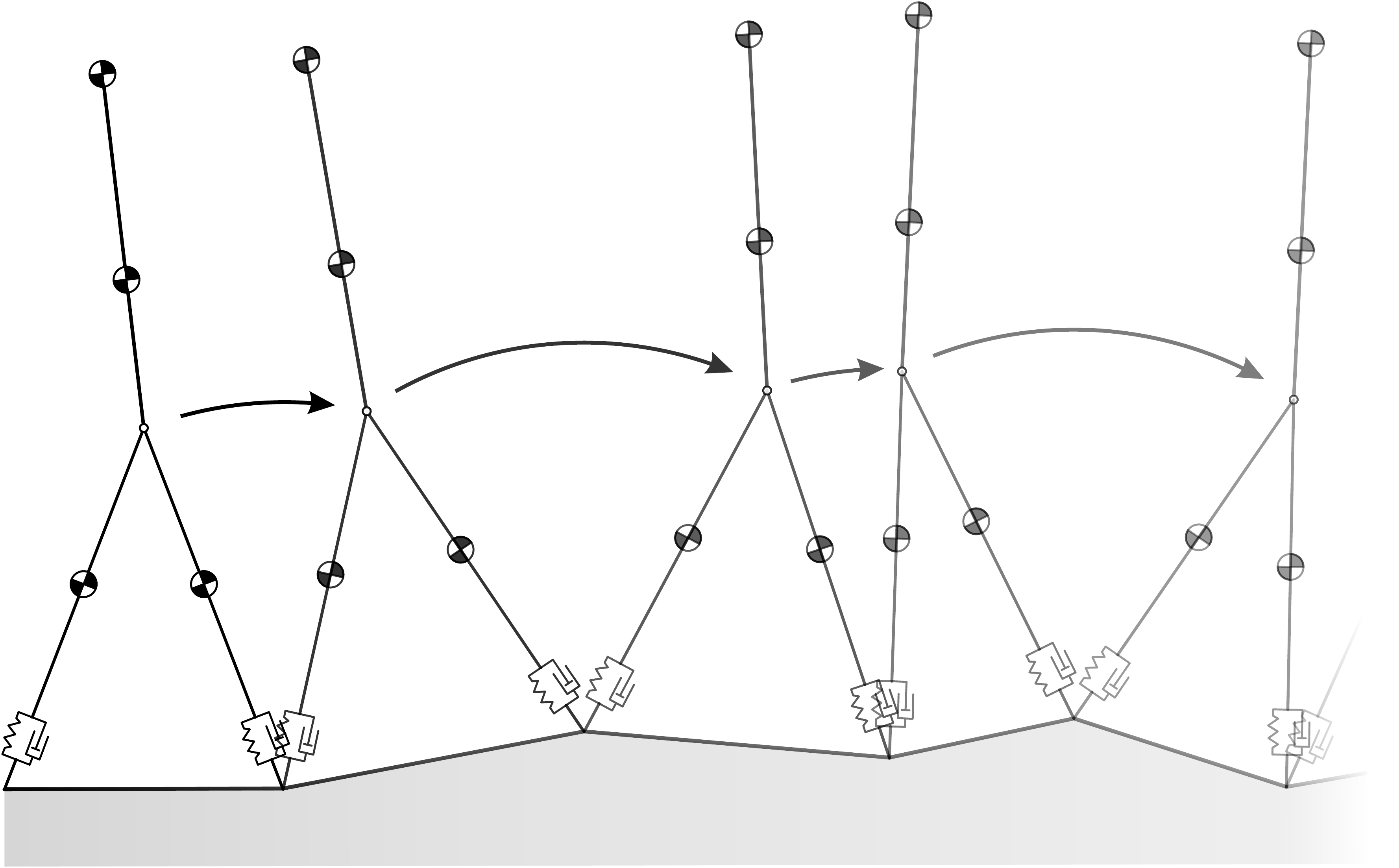}
\caption{\label{fig:roughTerrain} The model of ground texture for our walkers}
 \end{center}
 \end{figure}

In our former work, we simulated the walkers on a textured ground. For the case of our 2D walker, we model the uneven terrain by changing the slope of the ground at each step, following a centered Gaussian distribution (see Figure~\ref{fig:roughTerrain}). We can make the terrain more or less rough by changing the standard deviation (std) of the Gaussian law. For any non zero standard deviation, the probability to fall tends to 1 when time tends to infinity.

To compare robustness performances, we use a metrics adapted to textured terrain~\citep{byl2009ijrr}. This metrics called Mean First Passage Time (MFPT) is derived from the analysis of metastable systems. MFPT of walkers on stochastically rough terrain is the average number of steps the walker makes before falling. This metrics has the advantage of being easy to interpret and to give an idea on the real performances of a walking system.  Furthermore, it synthesizes all the dynamics of reachable state space on rough terrain. In another previous work, we designed an approach to compute this value for complex walking system~\citep{benallegue2013rss}.

\subsection{MFPT on textured ground} \label{subsec:resMFPT}
\begin{figure}
\begin{center}
\includegraphics[width=\columnwidth]{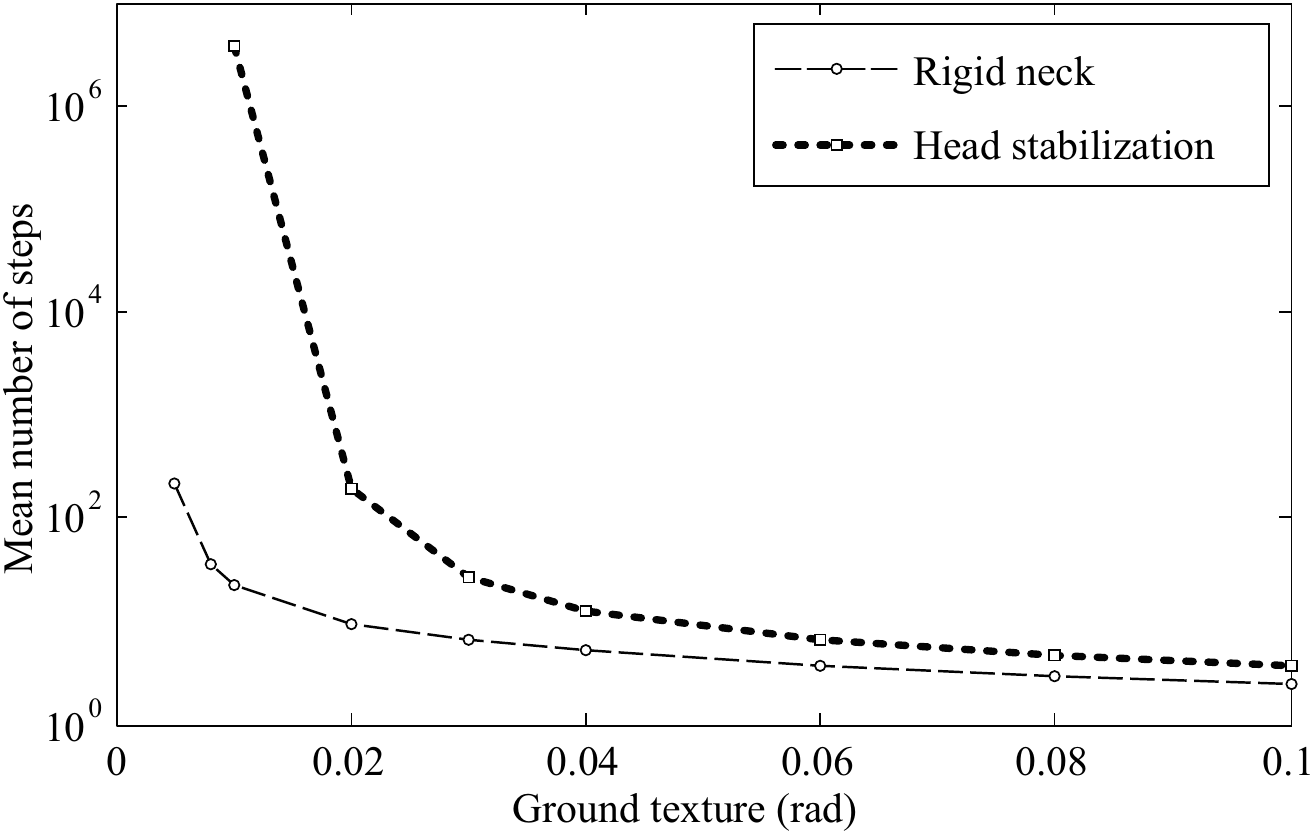}
\end{center}
\caption{ 
Mean number of steps of the walker models on different textures of the ground. By texture we mean the standard deviation of the ground slope. MFPTs are displayed in logarithmic scale. }
\label{PlotIdeal}
\end{figure}

On flat terrain and for both control models, it has not been possible to find an upper bound on MFPTs. However, as soon as a slight non-flat texture appears, the walkers' performances greatly differ from each other. In Fig.~\ref{PlotIdeal} we show the different values of MFPT obtained for different ground textures, compared between the head stabilization and the rigid neck models.

We see in the plot that head stabilization enables a substantial improvement of the mean number of steps on weakly textured grounds. The phenomenon can be seen from the example of 0.01 rad standard deviation. In this case, MFPT of the rigid neck model is 23 steps, while head stabilization guarantees MFPT of several million steps. This performance improvement persists as the ground texture increases, even if the difference declines. 

However, we don't believe this is the most appropriate way to interpret these results. Instead, we would emphasize more on the fact that the head stabilization moves the rigid-neck line of Fig.~\ref{PlotIdeal} to the right. That means that the same balance performances can be achieved on higher textures thanks to head stabilization. 

The next section introduces complementary results that give deeper insights into the mechanical contribution of head stabilization to walking dynamics.

\section{The effect of head stabilization on lower-body kinematics}\label{sec:kinematics}
The rigid neck Model A and head stabilization Model B have different kinematics and internal torques in the neck joints during walking. But in Model A, the head is attached to the torso which is stabilized upright as well with quite high gains. We may expect then that the difference between the two models would be very slight, especially regarding the fact that the head has a small mass compared to the rest of the body. However, as we have seen before, the difference between the models induces an important difference in terms of robustness to ground perturbations.

\begin{figure}
\begin{center}
\includegraphics[width=\columnwidth]{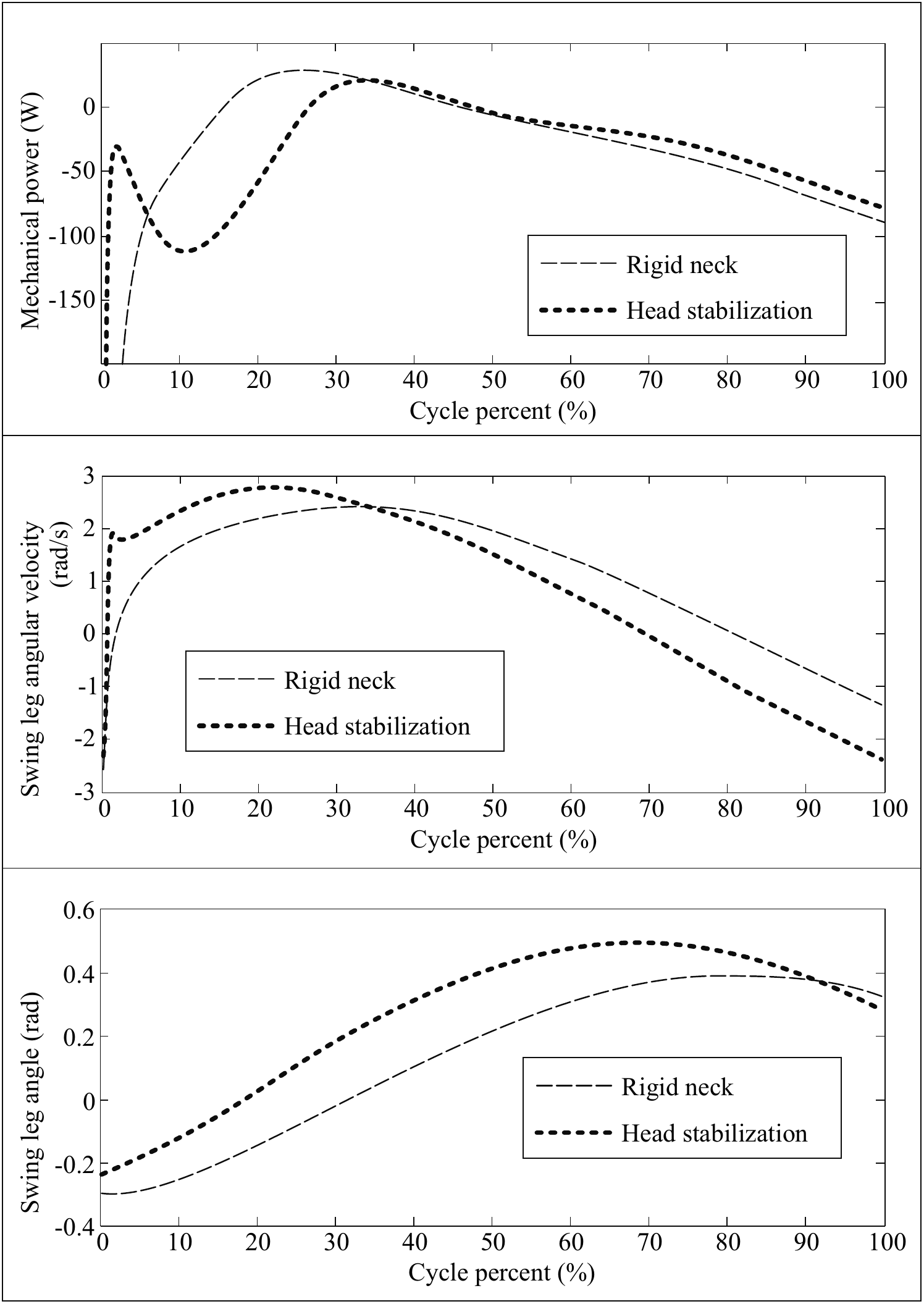}
\end{center}
\caption{ 
{ Mechanical power and leg velocity over the walking limit cycle.}  On the top: The mechanical power during phases of limit-cycle on flat ground. From 0\% to 30\% of the cycle, the distribution of mechanical energy consumption is different, the head stabilization offers a lower energy absorption following the impact. On the middle: The angular velocity of the swing leg during the limit-cycle. The influence of the mechanical power difference in the first 5\% is visible, the swing leg has a better initial velocity until 30\% of the cycle. On the bottom: The swing leg angle, we see that despite similar initial and final values, the angle is higher when the head is stabilized.
}
\label{MechanicalEnergyPlot}
\end{figure}

In order to remove all the disturbances and study the differences in ideal conditions, we study the dynamics when they are on the stable limit cycle. We start our dynamics analysis by studying mechanical energy during the gait. Here, by mechanical energy we mean the sum of kinetic and \emph{gravitational} potential energy. We do not include the potential energy stored in spring compression, or in deviation from the reference of PD controllers. We see in the top of Fig.~\ref{MechanicalEnergyPlot} a plot of the total mechanical power (time-derivative of mechanical energy) of both walkers during one cycle of walking.

At the beginning of the cycle, the walker is in a post-impact state. This part is truncated in the plot because of very low power values corresponding to the absorption of mechanical energy, mostly by the spring-damper at the toe. After the impact, the upper body is propelled forward by the conservation of momentum. The active trunk stabilization of both models dissipates a part of this motion to maintain the upper-body around upright orientation. This stabilization tends then to reduce the amount of the mechanical energy of the system, and thus has a negative power during almost all the cycle. Finally, following the hybrid nature of the system, at the pre-impact instant, there is a discontinuous impulsion given by the toe. This gives an `infinite' mechanical power at the last instant of the cycle. This value is obviously omitted in the plot. 

Since this motion is a periodic limit cycle, both mechanical powers have zero integral when we consider these truncated values. And even by subtracting the impulsion energy of the last instants the two plots have approximately the same negative integral.

In addition, what is striking about this plot is the difference that lies between the plots just after the impact. For instance, in the case of head stabilization, there is a noticeable amount of mechanical energy dissipation which is delayed from the very first post-impact instants (0 to 7\% of the cycle) to a later part of the cycle (7 to 30\% of the limit cycle). The origin of this difference is investigated in the following section on a simplified model of the upper body.

This phenomenon has an effect also on the lower body. The difference in mechanical energy at the beginning of the step involves partly the swing leg velocity. The plot on the middle of Figure~\ref{MechanicalEnergyPlot} shows the angular velocity of the swing leg during the limit cycle. There is a clear increase in swing leg velocity at the beginning of the step when the head is stabilized. The velocity decreases then later in the cycle. The effect appears at the bottom of Figure~\ref{MechanicalEnergyPlot} with different trajectories of the swing leg angle despite similar initial and final values. For instance during most of the limit cycle period the head stabilization guarantees higher swing leg angle.

This modification of the legs kinematics has an influence on the way the gait responds to ground perturbations. Indeed, having higher foot elevation enables to overcome more obstacles in the environment. But there is also another effect, which is to reduce forward falls: in the case of human gait, as for our simulated models, the imbalance caused by uneven and textured ground result mostly in \emph{forward} falls~\citep{Smeesters2001309}. A forward fall happens when the swing foot reaches the ground too close to the stance foot, and cannot compensate for the destabilizing linear and angular momenta after the impact. Therefore when the swing leg velocity is increased at the beginning of the cycle, the walker reaches earlier a balanced position with a forward leg better prepared for the next impact~\citep{wisse2007delft}. This is precisely what happens to our simulated model with a stabilized head. 

To understand the origins of this effect, we show in the next section how this difference in actuation affects a linearized model of the upper body alone.

\section{Linearized dynamics}
\label{sec:cartupperbody}
\subsection*{Cart upper-body model}

The upper-body, i.e. everything above the hips, represents the majority of the weight of our model, distributed on an upright structure which is broadly as tall as the legs. Therefore, during steady gait, it has a remarkable contribution to the dynamics of walking~\citep{benallegue2013icra}. Moreover, beside its simple presence, the internal dynamics of the upper-body has also its own effects on the gait, this is known for example for the case of articulated arms~\citep{Chevallereau2009itro}. Similarly, any difference in the dynamics of the head-neck system should have its effects on the whole-body motion. 

We consider here the upper-body model of our previous walker, which is the 2D chain composed with the three segments: the head, the neck, and the torso. This model is controlled similarly to the walker presented earlier with a rigid trunk system (Model A) and head-stabilizing one (Model B) and having the same control parameters as in~\citep{jpl2017ijrr}. Note that even if the models could remind of the classic inverted pendulum-like systems~\citep{wieber2015modeling}, with its unstable natural dynamics, the main difference is that all the rotational degrees of freedom of these models are actuated, giving then global stability of the control when coupled with the described control.

In the equilibrium state of Model B, the center of mass of each segment is vertically aligned with the previous and next joint. This system becomes therefore stiff regarding \emph{vertical} perturbation forces. Or in other words, the main difference that we would see between Model A and Model B will lie in the \emph{horizontal} component. For this reason, we put these models on simulated carts moving on horizontal planes. 

We define the state vector for the upper-body of Model B: $\xi_u=\begin{pmatrix}p & \dot{p} & \alpha & \dot{\alpha} & \gamma & \dot{\gamma} & \beta & \dot{\beta} \end{pmatrix}^\top$ where $p$ is the horizontal position of the cart. The state vector of Model A can be reduced by removing the four redundant last components.

\begin{figure}
\begin{center}
\includegraphics[width=0.6\columnwidth]{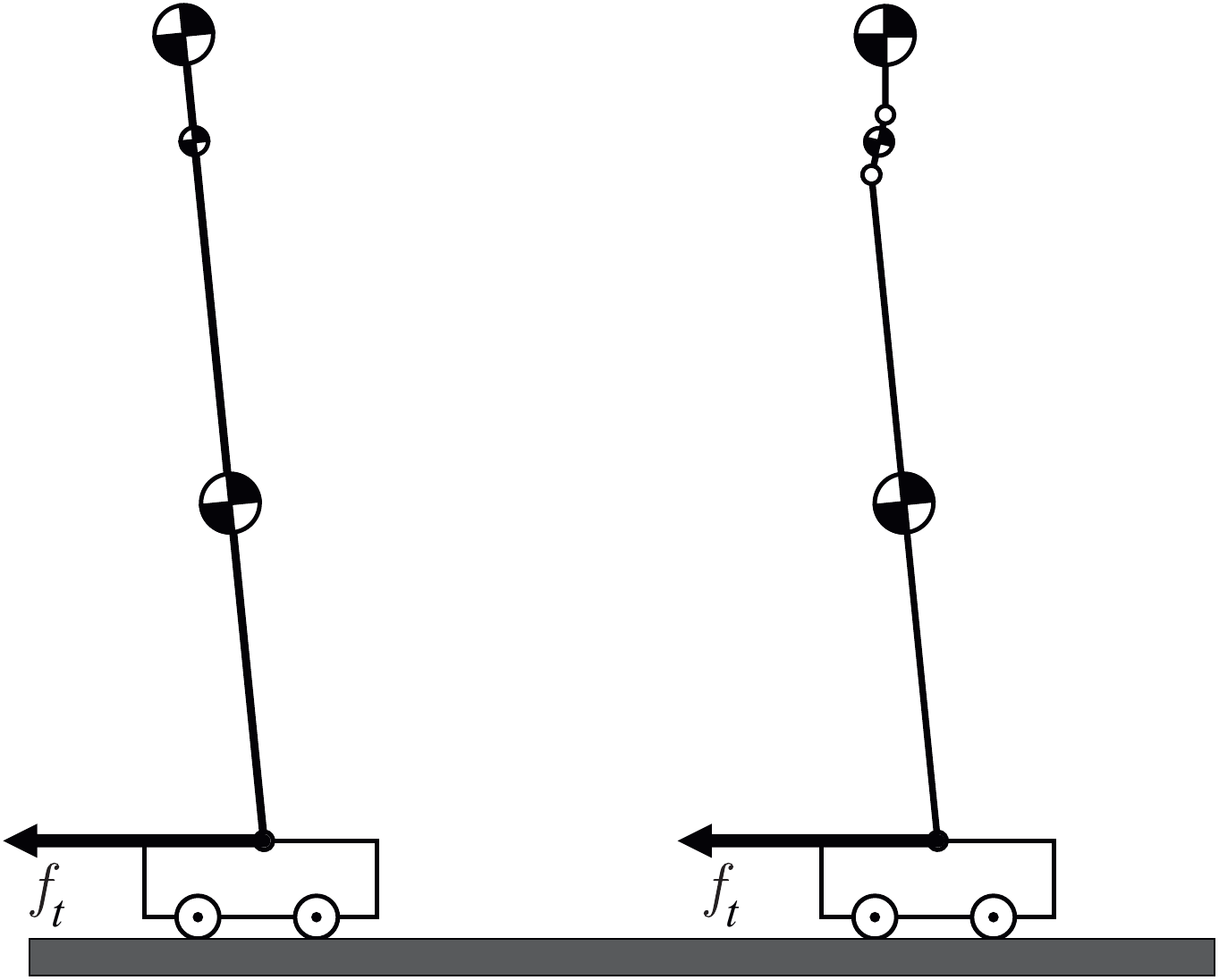}
\caption{\label{fig:upperbody-cart} The upper body models on carts. On the left the A model with a rigid neck and on the right the B model with an actuated articulated neck. The dynamics of these both models is linearized to be studied.}
 \end{center}
 \end{figure}

We equip the cart with an acceleration-driven actuator (see Figure~\ref{fig:upperbody-cart}). This ideal actuator generates perfectly the necessary force in order to track a reference acceleration $\ddot {p}$. This modeling choice allows comparing the dynamical responses in terms of reaction forces and energy. The dynamics can be written as
\begin{equation}
\dot{\xi}_u= f(\xi_u,\ddot {p}),
\end{equation}
where $f$ is the state dynamics which is different between Model A and Model B. Note that developing this dynamics boils down to solving a mixture between forward and inverse dynamics. That is because we know part of the forces through gravity and pure torque control law (equations (\ref{eq:pdtrunk})-(\ref{eq:pdhead}) in supplementary material) and part of the second-order kinematics through $\ddot {p}$.

The dynamics function $f$ is nonlinear because of trigonometric operators. We choose to linearize this dynamics around the equilibrium state ($\alpha=\beta=\gamma=\dot\alpha=\dot\beta=\dot\gamma=0$). The linearization enables us to catch the dynamics in the vicinity of this state while allowing easy frequency analysis and impulsion response study. The linearization provides us with the following approximated dynamics:
\begin{equation}
\dot{\xi_u}= A\xi_u+B\ddot {p}
\end{equation}
where $A$ and $B$ are matrices of appropriate dimensions. We can compute these matrices for both models by using available equations provided by Newton-Euler dynamics and kinematic constraints. We redirect to the supplementary material for detailed computation of these matrices.

\subsection{Spectral analysis}
\begin{figure}
\begin{center}
\includegraphics[width=\columnwidth]{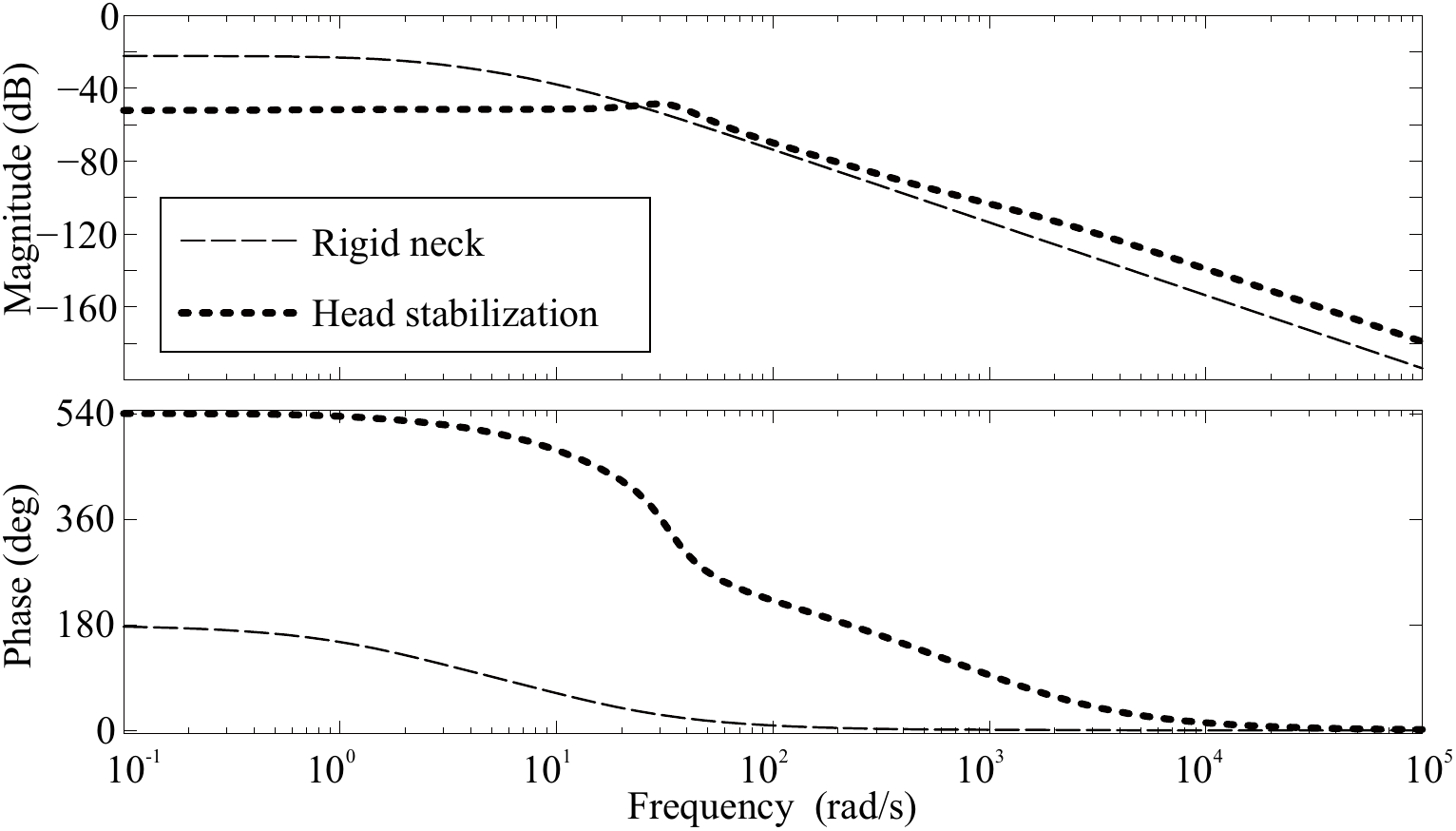}
\caption{\label{fig:headbodeplot}Bode plot of the response of $\beta$ the head angle to $\ddot p$ the cart acceleration inputs for the linearized cart-upper-body model with rigid trunk and head stabilization.}
 \end{center}
 \end{figure}

The first assessment that can be made is on the efficiency of the head stabilization. The head inclination $\beta$ is part of the state vector and can, therefore, be studied in the Bode diagram of Figure~\ref{fig:headbodeplot}. We can see that the head-neck PD control is more efficient than the rigid trunk to stabilize the head for frequencies lower than 30 rad/s (about 4.77Hz) which is sufficient to contain most of walking dynamics. The dynamics of head orientation is very comparable for higher frequencies, which may appear mostly during impacts. We can state then that the head is successfully stabilized thanks to our control model.

\begin{figure}
\begin{center}
\includegraphics[width=\columnwidth]{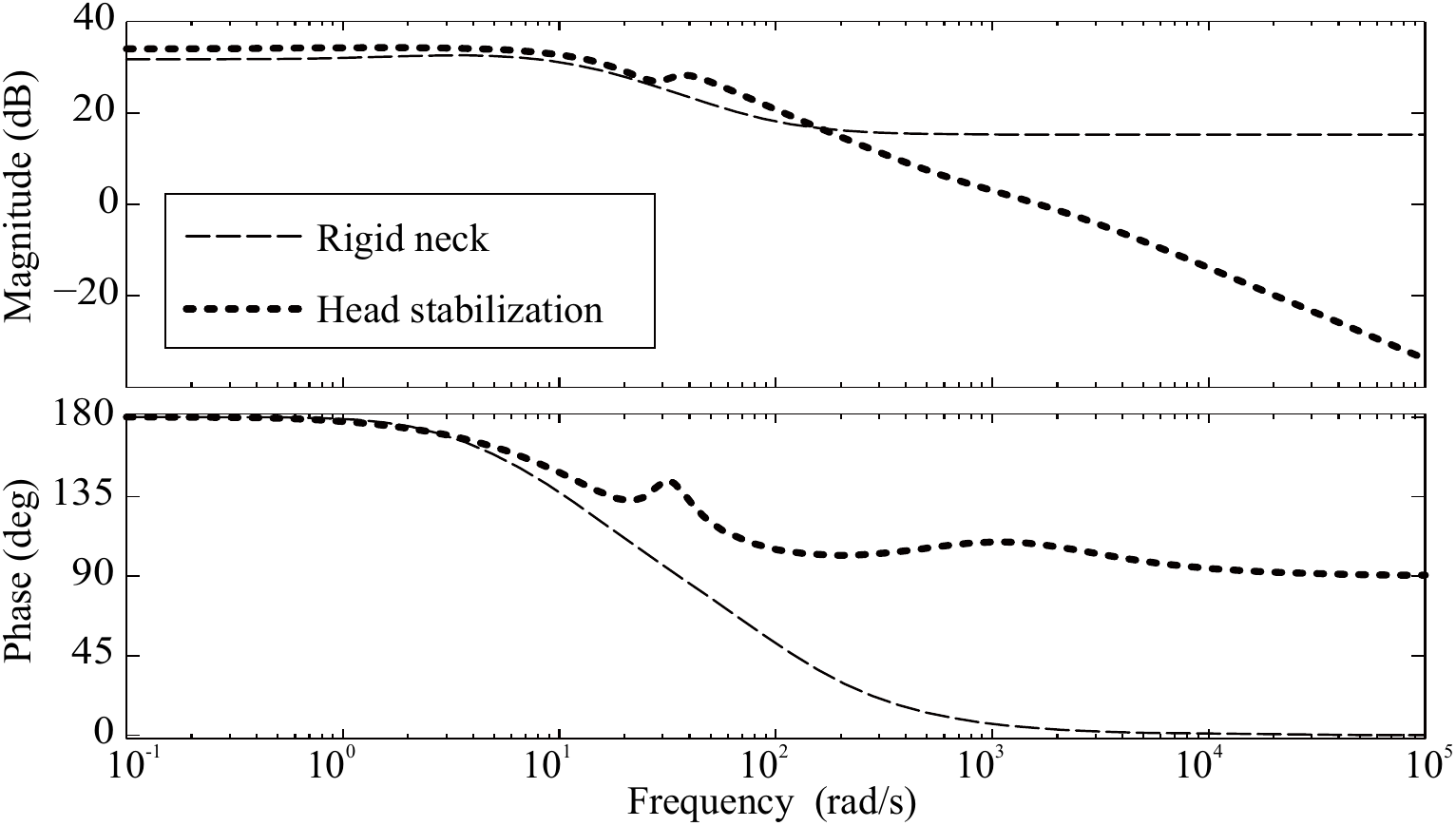}
\caption{\label{fig:forceplot}Bode plot of the response of the force $f_t$ to $\ddot p$ cart acceleration inputs for the linearized cart-upper-body model with rigid trunk and head stabilization.}
 \end{center}
 \end{figure}

More interestingly, the horizontal forces $f_t$ that are applied by the cart on the trunk are also linear outputs of the system. % Indeed, the forces are proportional to the accelerations of the center of mass which can be obtained linearly through the state $\xi_u$ and the actuation $\ddot p$ (see Appendix~\ref{sec:linearization} for details). 
This output represents the necessary force required to move the cart at a given acceleration $\ddot p$. Therefore, this response can be seen as the \emph{apparent inertia} of the upper-body seen as a black box. The Bode diagram of this output is presented in Figure~\ref{fig:forceplot}. 

We see in this plot that this apparent inertia decreases for Model B for high frequencies, compared to Model A, due to the mobility of its head-neck system. This modification of the dynamics is partly due to the fact that the moment of inertia of the rigid trunk regarding the hip joint is much higher than the sum of moments of inertia of each segment separately relative to the corresponding lower joint. Thanks to this property, the head-neck system appears to play the role of a \emph{low-pass filter} on the force dynamics of the upper-body.

This difference observed in the forces imply that the impact responses are different. Since impacts can be considered as acceleration Dirac functions they can be simulated on our linearized upper-body model.

\subsection{Response to impacts}
The impact during the limit-cycle of walker models happens just after the limit state $\xi_l$, and its dynamics is determined by the value of this state. This value is different between Model A and Model B but it generates roughly comparable post-impact dynamics. For instance, the outcoming horizontal motion of the hip joint is relatively similar between Model A and Model B and can be approximated by a discontinuous transition from 1.2 m/s velocity to 1 m/s. This transition can be simulated as an impulse response of this linear system. 

\begin{figure}
\begin{center}
\includegraphics[width=\columnwidth]{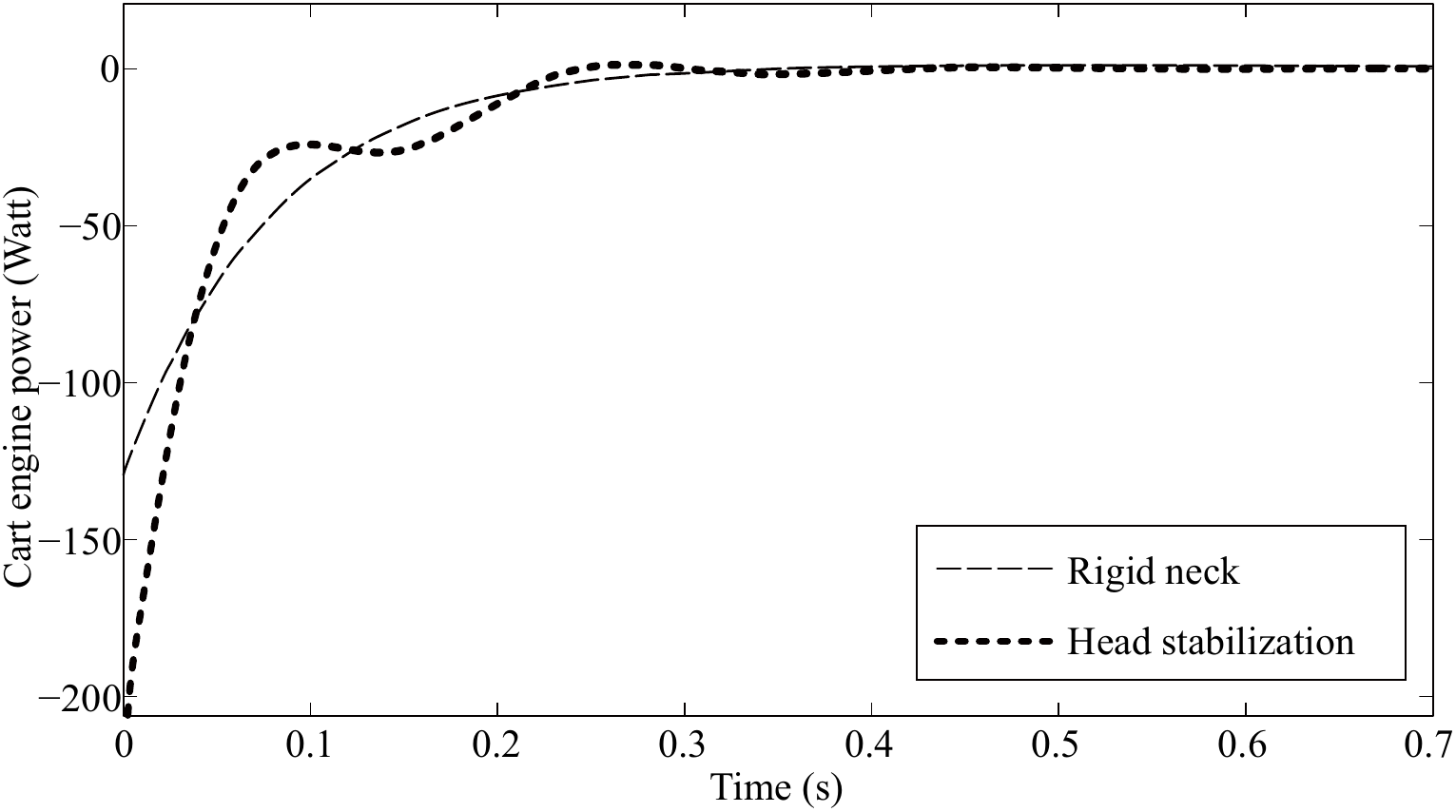}
\caption{\label{fig:linearcartpower}Time-plot of the impact response of the cart mechanical power for both models.}
 \end{center}
 \end{figure}

Similarly to the walker model, we show the response in terms of mechanical power as a time-plot in Figure~\ref{fig:linearcartpower}.

The integrals of the two curves have comparable values (-9.3~J for Model A and -10.3~J for Model B) since both models start and end at specific energy levels determined by the pre-impact and post-impact velocities. Also, we see in this plot that after 0.5 seconds, the cart has almost no need for power to keep the new constant velocity. However, before this instant, the distributions of this energy consumption are relatively different. Specifically, the head stabilizing Model B reaches a low level of energy consumption significantly faster than the rigid neck Model A but keeps this level longer (between 0.08~s to 0.15~s). Of course, this difference is due to the difference in terms of apparent inertia showed earlier. 

It is important to note that this plot looks very similar to the mechanical power at the top of Figure~\ref{MechanicalEnergyPlot}. This is because it describes the very same phenomenon which allowed our model to better overcome ground perturbations. The head-neck system acts as a low-pass filter on the dynamics of upper-body. In some way, this amounts to save a part of mechanical energy and to dissipate it at a more appropriate phase of the walking cycle. This allows propelling the leg forward to better prepare for the next step.

This dynamical effect obviously depends on the parameters of the controllers, specifically the actuator gains, and might have different outcomes if they had different values. In the following section, we study how the viability and the robustness are sensitive to various gain values.

\section{Sensitivity to control parameters}
The gait of our walkers is the result of the coordination between the natural dynamics of the mechanical rigid structure and the introduced forces into the system through active actuators at the joint level or the passively compliant toe. Nevertheless, this coordination emerges from simple control schemes and only a few parameters define the nature of their responses. To produce the robustness results seen in the previous section, these parameters did not need to be finely tuned. They were actually in the first set of values we tested that allowed the emergence of a stable gait for both of our walkers. However, not all values for these gains produce a viable walker, i.e. a walker able to have stable gait on flat ground. And the robustness also depends on these parameters to the same extents.

Therefore, we study, in this section, the sensitivity of our effect to variations of these control parameters, in terms of viability and robustness to ground perturbations. To do that, we generate random gain parameters for both walkers and we study these two controllers for every combination.

\subsection{Generating random parameters}
Both walkers share the same architecture up to the neck joint. They share then at least 7 control parameters: the stiffness and damping of the toe compliance, the impulse velocity reference and the proportional and derivative gains for the interleg angle and for the trunk inclination. If these values are too small, the walker will either generate less power than it dissipates or will be too compliant and unbalanced. If the values are too high, the robot would either be too stiff to allow a stable gait or would produce unstable simulations. 

For each of these control parameters, the value was generated uniformly ranging between the quarter and the double or the quadruple of their initial values. A dedicated section in the additional material provides details on this generation, and on the methods, we used to measure the viability and the robustness to ground perturbations.

\subsection{Viability}
\begin{figure}
\begin{center}
\includegraphics[width=\columnwidth]{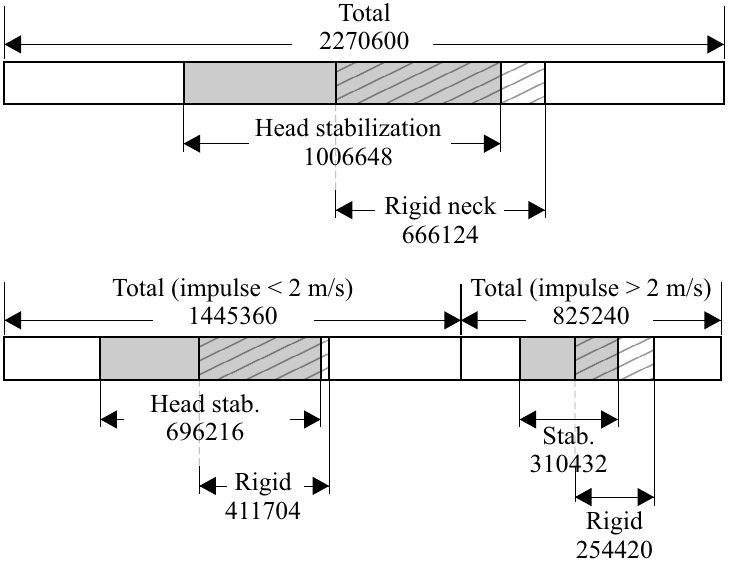}
\end{center}
\caption{ 
{ Representation of the viability tests on walkers with random parameters.} There were 7 different generated parameters, which cannot be represented in a plot. Therefore each bar represents the number of combinations lying in a set. The clear area represents the total sampled parameters, the solid gray area represents the viable parameters for head stabilization and the striped area represents the viable parameters for the rigid trunk. The representation allows seeing the overlap between viable rigid trunk walkers and viable head stabilization walkers. On the top are the global results on parameter sampling, and on the bottom are the same results split into stronger and weaker impulsions. 
}
\label{fig:randomize-viability}
\end{figure}

For each combination of the control parameter, we tested the viability of the walker, i.e. its ability to generate stable gaits on even flat ground. On a total of 2,270,600 combinations tested, 1,121,695 (49\%) were not viable for any walker, 523,867 (23\%) were viable for both walkers, 482,781 (21\%) were viable for the head stabilization walker and not viable for rigid neck and only 142,257 (6\%) were viable only for the rigid neck walker. See the top of Figure~\ref{fig:randomize-viability} for a more visual representation of these results. 

We call exclusive viability the situation in which, for the same parameters, one of the models was able to reach a steady gait and the other not. It is either due to the inability of one control to generate a stable motion while the other achieves it, or when the simulator was not able to start inside the basin of attraction of the failing walker but succeeded for the other one.

These results mean that the head stabilization allows more controllers to perform a viable gait than the rigid trunk walker, mostly due to better control and a larger basin of attraction, and only a few combinations are exclusively viable for the rigid trunk.

The parameters giving exclusive viability for the rigid neck tend to have a higher impulse velocity in average (2.31~m/s compared to 1.68~m/s for viable head stabilization), which is much more energy-consuming. This can be seen by constraining the impulse velocity to be below 2~m/s, which is twice the velocity of our original system. In that case, we have 1,445,360 combinations, among which 383,930 (26\%) are viable for both walkers, 312,286 (21\%) are viable exclusively for the head stabilization and 27,774 (2\%) only are viable for the rigid trunk and not for the head stabilization. This means that for low energy walk the head stabilization provided almost twice more viability than having a rigid trunk, without losing the viability of almost any control parameter.

Conversely, in the case where the impulse is bigger than 2m/s. Head stabilization had only 21\% of exclusive viability, which is only a bit more than 14\% of the exclusive viability of the rigid trunk. See the bottom of Figure~\ref{fig:randomize-viability} for a more visual display of these split results.

Another particular property of the combinations exclusively viable for rigid trunk is that they have lower stiffness and damping for trunk upright stabilization (385.49~Nm/rad and 131.85~Nms/Rad compared to 401.12~Nm/rad and 174.68~Nms/Rad) which makes it more prone to deviate from the limit cycle when subject to perturbations such as rough terrain.
Indeed, we next test the randomized parameters for these walkers when exposed to ground perturbations.

\subsection{Robustness to terrain disturbances}
In this simulation, the control parameters were generated following the same distribution and tested on a rough terrain having 0.03~rad of standard deviation. Here we separate also the parameters into two sets, one with impulse velocities higher than 2~m/s, the high energy group, and the other with weaker impulses, the low energy group.

The performance estimation is performed also using MFPT metrics. But due to the required time to obtain this measurement, the number of tested configurations is much smaller than for the simple viability analysis.

\begin{figure}
\begin{center}
\includegraphics[width=\columnwidth]{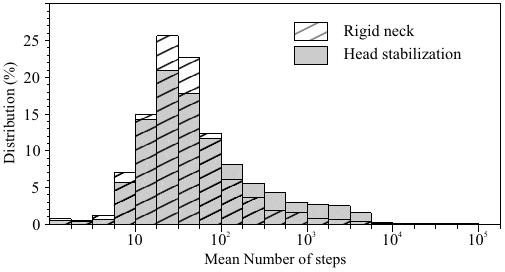}
\caption{\label{fig:randomizedDensity} The distribution of the mean number of steps when generating random control parameters with weak impulsion ($<$2~m/s). The MFPTs are shown in a logarithmic scale with each tick is $\sqrt[4]{10}\simeq 1.78$ times higher than the previous one. }
 \end{center}
 \end{figure}

For the lower energy group, there were 7713 configurations which were viable for both walkers. For these configurations, the simulations start at the limit cycle and the comparison between the MFPTs of these walkers is only a matter of robustness to ground perturbations. Among these values, 4516 (58\%) had higher MFPTs for head stabilization than for rigid trunk, and the remaining had higher MFPT for rigid neck system. This can seem a bit balanced between the walkers, however, the average improvement when head stabilization is better is of 1256 steps, while it is only 269 steps when the rigid trunk is better. Indeed, the average number of steps for the rigid neck is 167 while for head stabilization it is 754 steps. To emphasize this difference, the distribution of these MFPTs is presented in Fig~\ref{fig:randomizedDensity}, and shows that head stabilization provides higher robustness to ground perturbations. Head stabilization seems to provide a strong advantage for the robustness of the walker with regard to ground perturbation.

 However, this property is completely reversed for the set of parameters with stronger impulsions ($>$2 m/s). Head stabilization decreases drastically the robustness to ground perturbations. The average number of steps for the rigid neck is of 10537 steps while it is only 222 steps for head stabilization. This result may have two reasons: the first one is that the impulsion alone provides the swing leg with a high initial velocity, which makes the contribution of the head stabilization much less relevant, and the second one is that the high velocities of the resulting gait create large motions of the head which disturb the walking trajectories. It could be hypothesized that the quality of the resulting gait could be improved by using a stiffer head stabilization. However, going to stiffer actuation decreases the efficiency and the ``passivity'' of the gait. More than that, highly dynamic motion create large impacts that are not properly simulated with this kind of dynamic engines.

Finally, the head stabilization effect of gait seems to be related a lot on the need for the walker to delay the response to the impact to help the swing leg to overcome obstacles with minimum additional energy. These considerations, together with the impact and extensions of this result are presented in the following conclusive discussion.

\section{Discussion and Conclusion}\label{sec:discussion}
In this paper, we have investigated the origins of the effects of head stabilization on the dynamics of the walking dynamics in general and on the upper-body dynamics in particular. The study has been performed on simple 2D models, in the sagittal plane, aiming at mimicking the mechanical features of human gait while minimizing its complexity and the variety of parameters defining its dynamics. Therefore, despite the anthropometric values used, we intend to extract from this model more qualitative observations than quantitative measurements of the dynamics.

The head stabilization appears to provide a substantial dynamical advantage over the model with a rigid neck. This effect enabled the walker to generate intrinsically more stable gait kinematics and to obtain significantly better performances in terms of robustness to environment perturbations.

In a deeper investigation, we studied how different gaits produced by various control parameters are influenced by head stabilization, we have shown that this effect is rather generalizable as far as the walk is power efficient, but the effect becomes actually negative for high energy, highly dynamic gaits.

While this effect is significant for our model, we do not claim that the same phenomenon happens also for humans. We do not say that humans fall after few steps if they have their neck locked. That observation could transfer to humans by hypothesizing that the set of textures humans can absorb using their energy-efficient steady gait is extended thanks to the mobility of their vertebral column, especially cervical joints, and their active head stabilization. If the neck joint is locked, instead of falling, humans compensate for this loss and move to stiffer and costlier actuation to keep balance.   

In fact, our results fit with clinical observations on humans. The unsteadiness and the loss of
balance resulting from head-neck system sensorimotor disturbances have been widely documented. 
Deficiency in neck-joint motor abilities
have an important impact on balance, whether in chronic neck injuries~\citep{Stokell2011mt} or in one-time
experimental induced impairment~\citep{bove2002jn,Vuillerme2005nl}. It has even been suggested that the impairments
in the neck somatosensory inputs and sensorimotor control are as important for balance as a
lower-limb proprioception loss following a knee or an ankle injury~\citep{Treleaven2008mt}. Therefore, our study opens potentially new clinical perspectives in diagnosis and therapy. However, it is necessary to remind that in the case of humans the sensory system is also involved, and it is difficult to separate it from the pure mechanical effect of head stabilization. We think that our work could give a new perspective aiming at disentangling the mechanics from the control and helping to devise clinical protocols to detect impairments and improve recovery. Nevertheless, we note that the entanglement of the sensory-motor system in head stabilization, especially during locomotion, can be not only a tool to detect impairnesses~\citep{schubert2002vertical} but also a means to help rehabilitation patients with vestibular deficiencies~\citep{herdman2014vestibular}, where the gait is considered as a task allowing to assess the resulting recovery~\citep{Herdman2003jama,schubert2008mechanism}.   

%Therefore, no conclusions can be drawn on humans without dedicated experimental research on human subjects.

In addition, we believe our results may give an insight into the role of the control of a vertebral column in the management of gait mechanics. The whole upper-body inertia and elasticity could similarly store energy and use it at the relevant instant. This can inspire future mechanical designs of humanoid robots in order to include vertebral-like structures~\citep{ly2011iros,takuma2018auro}.
 
% It is worth to emphasize that the importance of this mechanical effect is also due to the weakness of lower limbs actuation, which makes the dynamics more sensitive to external forces. Indeed, other stiffer controls can be found for which the mechanical contribution of the head stabilization is reduced. However, stiff controllers are energy inefficient compared to steady gait where the walkers take advantage of the natural passive dynamics of the body. On the contrary, stiff actuation and high-frequency control are required to guarantee balance for highly uneven and unpredictable environments, and in this case, head stabilization is not expected to play such an important mechanical role.

Of course, one could design a different control of other parts of the body, such as the legs, and reaches the same results if not better. But our study focuses on the specific issue of exploring, at least qualitatively, the mechanical effect of head stabilization on the dynamics of gait, which is a recognized feature of human walking. On the contrary, the design and control of optimal models is the topic of our related research regarding simultaneous model design and control of robots in the same optimization loop~\citep{saurel2016simpar}.

Finally, an interpretation of our study is that head stabilization may be a \emph{heuristic} answer to the question of taking advantage of head mobility during walking.
Indeed, head stabilization it is likely \emph{not} the optimal control of the neck regarding balance. Nevertheless, it could be a very simple control that produces a complex behavior with significant benefits. 

In that perspective, the expected outcomes of this study will be to consider the design of future humanoid robots by introducing different mechatronic architectures and new sensory-motor systems.

%\begin{verbatim}
%\begin{biog}
%To typeset an
%  "Author biography" section.
%\end{biog}
%\end{verbatim}

%\begin{verbatim}
%\begin{biogs}
%To typeset an
%  "Author Biographies" section.
%\end{biogs}
%\end{verbatim}

%\newpage

\begin{appendix}
\section*{Supplementary Material}

In this supplemental material, we present the control and the actuation of the two models, then how the dynamics is linearized for the cart-upper body model, we describe how the mean first passage time was computed and finally, we give details on the generation and tests of random control parameters.

\section{Actuation Models}\label{sec:actuation}
\subsection{Toes}

The exchange between the swing phase and the stance phase occurs at impacts of the swing leg with the ground. Impacts are considered inelastic and contacts are considered perfect with no slipping. The toe of the stance leg has a spring-damper dynamics. The contact force follows the direction of the stance leg and its magnitude has a proportional-derivative (PD) expression:
\begin{equation}
f_t=-K_{toe,p}(l_p-l_{p,0})-K_{toe,d}\dot{l_p}
\end{equation}
where $K_{toe,p}=50000$~N/m is the elasticity of the spring and $K_{toe,d}=2000$~N$\,$s/m is the damping factor. This force is applied only when it is positive because of the unilateral force constraint of the contact (the ground cannot pull the body).

When a leg is in a swing phase, its toe comes back instantly to the rest position $l_{p,0}$ of the spring, and remains constant until the end of the swing phase. We denote then simply by $l_p$ the length of the stance leg.

\begin{figure}[]
\begin{center}
\includegraphics[width=0.8\linewidth]{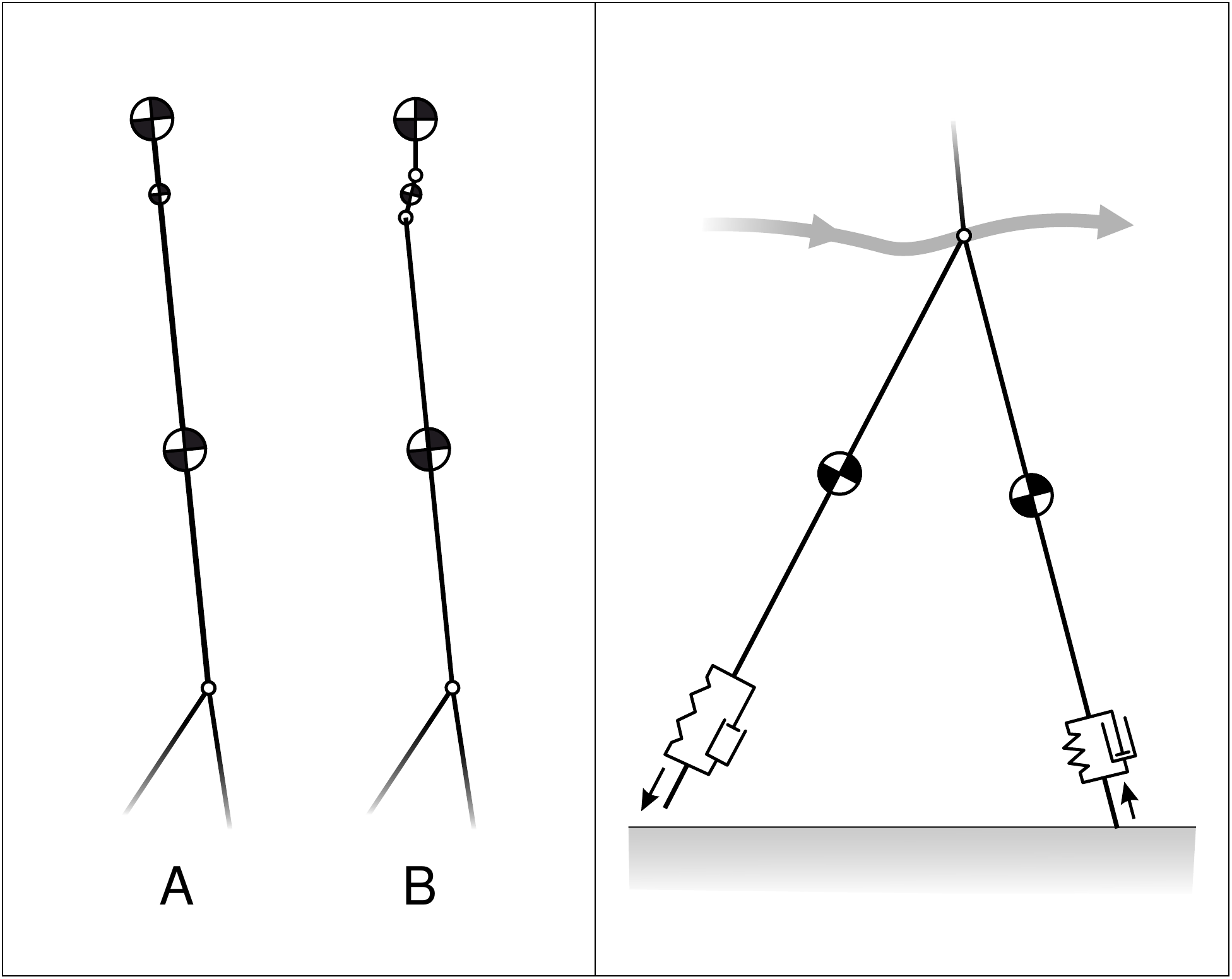}
\end{center}
\caption{ 
Actuation model.  On the left, we see the two models of upper-body (for Model A and Model B).  On the right, a scheme of the spring-damper at impact and the impulsion at take-off.
}
\label{Stabilization} \label{lowerBodyActuation}
\end{figure}

The walkers lose a part of their mechanical energy at each impact. They require then to be actively fed with an equivalent source of energy. Therefore, at the instant of the take-off of the stance leg, a velocity controlled impulsion is applied to the ground to give propulsion to the robot (Figure~\ref{lowerBodyActuation}). The required force for this impulsion is $f_t$:
 \begin{equation}
f_t=h(\dot{l}_{p,r})
\end{equation} 
where $\dot{l}_{p,r}=1$ m.s$^{-1}$ is the desired velocity and $h$ is the controller function. Most modern dynamic simulators provide implementations of ideal velocity-driven motors. These can be seen as models of velocity-driven motors with short response time and tracking high quality enabling the motor to reach desired velocities within one iteration of the dynamic simulator. In reality, the simulator solves the problem of finding the exact force that has to be applied to the joint during one time-sample to reach this velocity. This gives an automatic computation of $h$ which has no closed-form. We use it then to apply the force $f_t$ during one time-step of simulation.

 The idea of velocity-driven impulsion is to compensate for the energy loss of foot impact. Moreover, by having a constant toe-off velocity at the beginning of each step, we ensure to start every step with comparable levels of energy and momentum. Therefore the impulsion force is smaller for high-velocity stepping and bigger for low-velocity ones, and this helps to reject a part of the deviations from the limit cycle.
 
 In addition, even if the role of propulsion during human gait is still a subject of debate, several studies show the importance of calf and ankle in the propulsion during the support phase~\citep{hill1953prsl,winter1983energy}, their contribution to swing leg initiation~\citep{hof1992calf,meinders1998role} and their involvement in the control of trunk dynamics~\citep{neptune2001contributions}.

\paragraph{Interleg joint}
 The inter-leg joint is controlled by a PD pure torque generator toward a reference angle:
 \begin{equation}
\tau_{hip}=-K_{hip,p}(\theta-\theta_r)-K_{hip,d}\dot{\theta}
\end{equation}
where $K_{hip,p}=10$~N$\,$m/rad is the proportional gain, $\theta_r=0.3$~rad is the reference angle and $K_{hip,d}=1.5$~N$\,$m$\,$s/rad is the derivative gain. We see that these values are small, even when they are scaled to natural human mass, in order to reduce energy consumption and preserve the natural dynamics of the legs.

\paragraph{Trunk}

The trunk stabilization is achieved by applying torque on the stance leg similarly to what is done in~\citep{McGeer1990ijrr}. 
The trunk torque is given by: 
 \begin{equation}
\tau_{t}=-K_{t,p}\alpha-K_{t,d}\dot{\alpha}, \label{eq:pdtrunk}
\end{equation}
where $\tau_{t}$ is the trunk to stance-leg torque, $K_{t,p}=300$~N$\,$m/rad is the proportional gain and $K_{t,d}=150$~N$\,$m$\,$s/rad is the derivative gain.

\paragraph{Head stabilization}
For Model B, there are two other controllers, which are the neck stabilization and head stabilization, they are also controlled by PD pure torque generators. Their torques expressions are the following:
 \begin{align}
\tau_{n}&=-K_{n,p}\gamma-K_{n,d}\dot{\gamma}\label{eq:pdneck}\\
\tau_{h}&=-K_{h,p}\beta-K_{h,d}\dot{\beta}\label{eq:pdhead}
\end{align}
where $\tau_{n}$ is the torque applied to the torso-neck joint, $K_{n,p}=50$~N$\,$m/rad is the neck proportional gain, $K_{n,d}=0.6$~N$\,$m$\,$s/rad is the neck derivative gain, $\tau_{h}$ is the torque applied to the neck-head joint, $K_{h,p}=150$~N$\,$m/rad is the head proportional gain and $K_{h,d}=1$~N$\,$m$\,$s/rad is the head derivative gain.

\subsection{Other details on the dynamics} Some additional details have to be specified hereinafter.

A fall is spotted by detecting contact between the floor and a non-support limb. 

Since the walker has no knees, The stance leg is a little bit shorter than the swing leg because of the spring of the stance being compressed under the effect of the weight of the walker. Therefore when the legs are parallel, the swing leg touches the ground, even during the flight phase of the walker. This would stop any gait from being performed. This problem is classically taken into account in simulation by neglecting the contact until a threshold angle between the legs is reached, and this is how we proceeded. Beyond the threshold, the collision detection is activated to catch the actual impact with the ground. In actual compass walkers, a usual way to avoid it is to rock the walker sideways in order to clear the swing trajectory, but for 2D constrained walkers, the solution is to use stepping tiles to ensure the swing leg does not touch the ground~\citep{McGeer1990ijrr}.  

 At the impact of step $k$, there is a pre-impact state $\xi(t_k)^-$ and a post-impact state $\xi(t_k)^+$ where $t_k$ is the time instant of the $k$-th impact. The set of all possible pre-impact states constitute a manifold that is called Poincar\'e section $\mathcal{S}^*\subset \mathcal{S}$. Therefore at the end of each step, the state of the robot $\xi(t)$ reaches $\mathcal{S}^*$. The process of returning to this manifold is called a first recurrence map or a Poincar\'e map. The number of steps is then the number of times the walker's state reached $\mathcal{S}^*$ before to finish in a ``fallen state'' denoted $\xi_f$ which gathers all the states considered as ``fallen walker''. 
 
 In the presence of a limit-cycle $\mathcal{C}\subset \mathcal{S}$, we have one state value $\xi_l$ defined by $\mathcal{C}\cap\mathcal{S}^*=\{\xi_l\}$ such that on flat ground $\xi_l$ is the fixed point of the Poincar\'e map, i.e. $(\xi(t_k)^-=\xi_l) \Rightarrow (\xi(t_{k+1})^-=\xi_l)$. Furthermore, if the limit cycle is attractive, then for any neighborhood $v(\xi_l)\subset \mathcal{S}^*$, and starting from any state $x_0$ in the basin of attraction $\bar{\mathcal{S}}$, there is a number of steps $k$ after which all the pre-impact states $\xi(t_{k+i})^-$, with $i>0$, lie inside the neighborhood $v(\xi_l)$.

\section{Mean First Passage Time Computation Algorithm}\label{annex:mfpt}
This pseudo-code of Algorithm~\ref{algo:mfpt} describes in details the algorithm summarized in the main text.
\begin{algorithm}
\caption{MFPT limit-cycle-based estimation}\label{algo:mfpt}
\begin{algorithmic}
\Procedure{MFPT}{$n$: \# of samples, $\sigma$: slope std}
\State $n_l, n_f, m_l, m_f \gets 0$
\For{$i \in \{1,\cdots,n\}$} \algcomment{sampling big loop}
\State $\xi \gets \xi_l$ \algcomment{initialize at limit-cycle}
\State continue $\gets$ \textbf{true}
\State $m \gets 0$ \algcomment{number of steps of the sample}
\Repeat \algcomment{one sample simulation}
\State slope $\gets$ SimulateGaussian(0,$\sigma$)
\State $\xi \gets$ SimulateStep($\xi$,slope)
\If {$\xi = \xi_f$ }  \algcomment{fall detected}
\State continue $\gets$ \textbf{false}
\State $n_f \gets n_f+1$ \algcomment{fall counter}
\State $m_f \gets m$ \algcomment{falling step-counter}
\Else 
\State $m \gets m+1$ \algcomment{successful step}
 \If {$\xi \in v(\xi_l)$ } \algcomment{limit-cycle return}
 \State continue $\gets$ \textbf{false}
 \State $n_l \gets n_l+1$ \algcomment{returns counter}
\State $m_l \gets m$ \algcomment{return step-counter}
\EndIf
\State \textbf{endif}
\EndIf
\State \textbf{endif}
\Until{continue = \textbf{false}}
\EndFor
\State \textbf{endfor}
\State $p_f \gets \frac{m_f}{m_f+m_s}$ \algcomment{probability of a sample to fall}
\State $r \gets \frac{1-p_f}{p_f}$ \algcomment{average returns number before to fall}
\State $\bar{n}_f \gets \frac{n_f}{m_f}$ \algcomment{average step number before falling}
\State $\bar{n}_l \gets \frac{n_l}{m_l}$ \algcomment{average step number before returning}
\State $\bar{m} \gets r\ \bar{n}_l+ \bar{n}_f$ \algcomment{MFPT}
\State \textbf{return} $\bar{m}$
\EndProcedure
\end{algorithmic}
\end{algorithm}

The neighborhood $v(\xi_l)$, also called limit kernel, is approximated by a multidimensional ellipsoid such that:
\begin{equation}
\xi \in X_l \Longleftrightarrow (\xi-\tilde{\xi}_i)^\top C (\xi-\tilde{\xi}_i)<d_0
\end{equation}
where $C=(cov\{\tilde{\xi}_i\})^+$ is the semi positive definite matrix defined by the Moore-Penrose pseudoinverse of the covariance matrix of the set $\{\tilde{\xi}_i\}$, and $d_0$ is a threshold defining the size of $v(\xi_l)$. For our case, we take $d_0=1000$.

\section{Linearized dynamics of cart-upper body model}\label{sec:linearization}
\subsection{The segmented linearized model}
\begin{figure}
\begin{center}
\includegraphics[width=0.5\columnwidth]{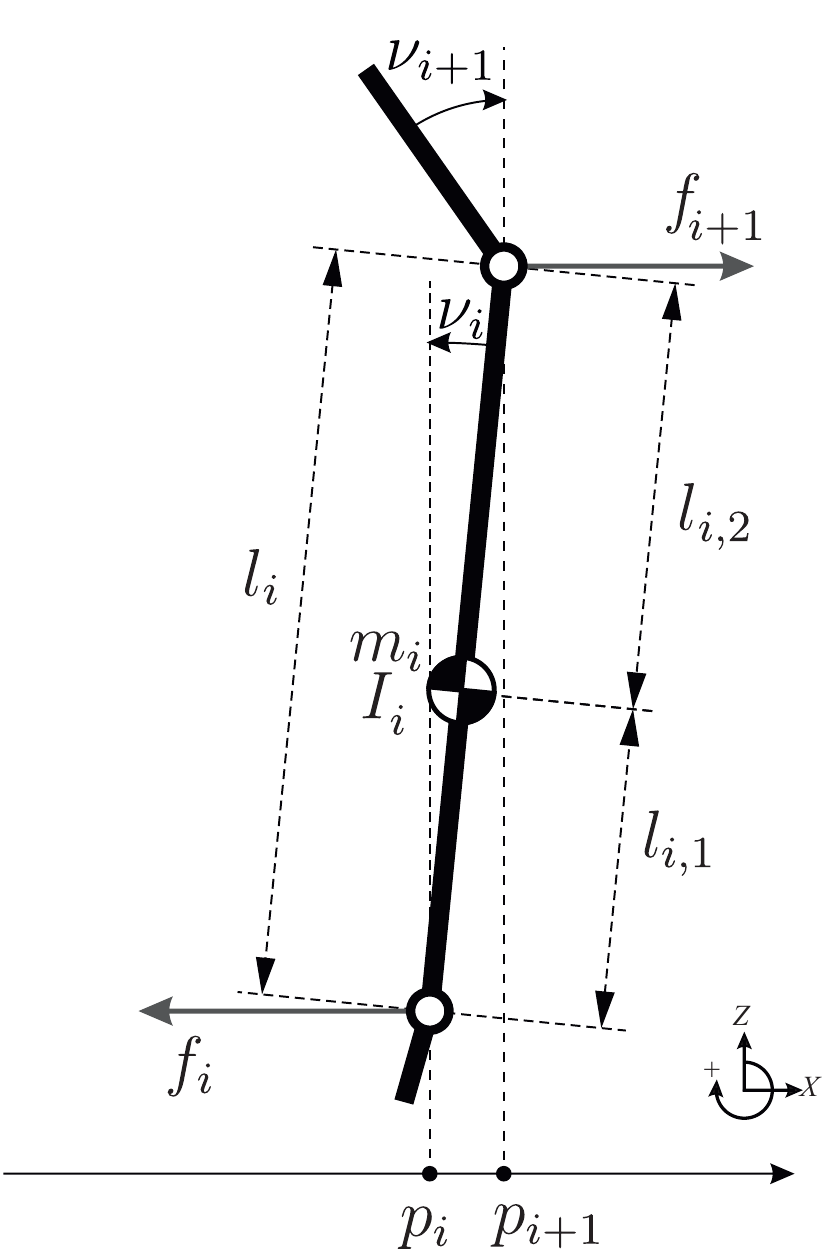}
\end{center}
\caption{ 
One segment $s_i$ of the model. }
\label{fig:segment}
\end{figure}

The linearized dynamics of the cart-upper body model can be obtained by writing all the equations provided by Newton-Euler dynamics and kinematic constraints. These equations can be described by considering each segment $s_i$ attached at the bottom and at the top to other segments $s_{i-1}$ and $s_{i+1}$ respectively (see Figure~\ref{fig:segment}). Each segment is subject to control torques applied at the bottom and top joints, it undergoes also forces applied by the upper and the lower segments. The kinematic constraint ensures obviously that the segments are linked together.

The equilibrium state is the state where the segment angles and their time-derivatives are null ($\nu_i=\dot{\nu}_i=0$ for all the values of $i$). Since the dynamics is linearized around the equilibrium state, this simplifies the dynamics by providing that the joints have constant height and only move in horizontal with $p_i$ is the position of the lower link of the $i$-th segment. This implies that the vertical force acting on each joint is the opposite of the weight of all above segments, leaving only horizontal forces $f_i$ to be determined. The linearization allows also approximating $\sin {\nu_i}$ by $\nu_i$. This enables us to produce the following equations:
\begin{align}
&\scalemath{0.85}{0=-K_{i,p}\nu_i-K_{i,d}\dot{\nu}_i+K_{i+1,p}\nu_{i+1}+K_{i+1,d}\dot{\nu}_{i+1}}\nonumber\\
&\quad\scalemath{0.85}{-I_i \ddot{\nu}_i -l_{i,1} f_i-l_{i,2}f_{i+1}+g(l_{i,1} m_i+l_{i}\textstyle \sum_{j=i+1}^{n}m_j)\nu_i\label{eq:lineuler}}\\
&\scalemath{0.85}{0=m_i (\ddot{p_i}+l_{i,1}\ddot{\nu}_i)-f_{i}+f_{i+1}\label{eq:linnewton}}\\
&\scalemath{0.85}{0=-\ddot{p}_{i+1}+l_i\ddot{\nu}_i+\ddot{p}_i\label{eq:linkinematics}}
\end{align}
where $K_{i,p}$ and $K_{i+1,p}$ ($K_{i,d}$ and $K_{i+1,d}$) are the proportional (derivative) gains of the stabilization of the current segment $i$ and the upper segment $i+1$. $m_i$ is the mass and $I_i$ is the moment of inertia around the center of mass for the segment $i$. $l_i$, $l_{i,1}$ and $l_{i,2}$ are for  the $i$-th segment the total length, the length of the lower part (between the lower link and the CoM) and the length of the upper part respectively. 

Equation~(\ref{eq:lineuler}) is Euler's relation for the variation of angular momentum around the center of mass of the segment $i$. Equation~(\ref{eq:linnewton}) is Newton's law for the variation of linear momentum along the horizontal axis. Equation~(\ref{eq:linkinematics}) describes the kinematics constraint linking each segment with the upper one, this equation does not apply if there is no upper link (at the end of the kinematic chain).

In the case of our upper-body models. We have two different controls, the model with head stabilization and the model with rigid neck. Let's study the linearization of each related dynamics:

\subsection{Stabilized head}
the equations of all the segments can be summarized as follows:
\begin{align}
Lz+E\xi_u+F\ddot{p}=0 
\end{align}
where  $\xi_u=\begin{pmatrix}p & \dot{p} & \alpha & \dot{\alpha} & \gamma & \dot{\gamma} & \beta & \dot{\beta} \end{pmatrix}^\top$ is the state vector described in the main text, $z = \begin{pmatrix} \ddot{\alpha} & f_t & \ddot{\gamma} & \ddot{p}_1 & f_n & \ddot{\beta} & \ddot{p}_2 & f_h \end{pmatrix}^\top$,  with $f_t$, $f_n$ and $f_h$ the internal horizontal forces applied at the hip-trunk, the trunk-neck and the neck-head joints respectively.
\begin{equation}
\scalemath{0.86}{
L=\begin{bmatrix}
0 & - l_{t,1} & 0 & 0 & -l_{t,2} & 0 & 0 & 0 \\
m_t l_{t,1} & -1 & 0 & 0 & 1 & 0 & 0 & 0\\ 
l_t&  0   &  0  &  -1 & 0 & 0 & 0 & 0\\
0 &  0 &  0 & 0& -l_{n,1} & 0& 0&-l_{n,2} \\
0 &  0 &  m_n l_{n,1}& m_n & -1 & 0 & 0 & 1\\
0 &  0 & l_{n} & 1 & 0 & 0 & -1 & 0\\
0 &  0 & 0 & 0 & 0 & 0 & 0 & -l_{h}\\
0 &  0 & 0 & 0 & 0 & m_hl_{h} & m_h & -1\\
\end{bmatrix}
},
\end{equation}
\begin{equation}
F=\begin{bmatrix}
0&
      m_t&
       1&
       0&
       0&
       0&
       0&
       0
\end{bmatrix}^\top,
\end{equation}
and
\begin{equation}
\scalemath{0.92}{
E = 
\begin{bmatrix}
0& 0& E_{0,2}& -K_{t,d}&                  K_{n,p}&  K_{n,d}&             0&   0\\
0& 0&                                  0&   0&                               0&      0&             0&   0\\
0& 0&                                  0&   0&                               0&      0&             0&   0\\
0& 0&                                  0&   0&   E_{3,4} &    -K_{n,d}&             K_{h,p}&   K_{h,d}\\
0& 0&                                  0&   0&                               0&      0&             0&   0\\
0& 0&                                  0&   0&                               0&      0&             0&   0\\
0& 0&                                  0&   0&                               0&      0&  E_{6,6}& -K_{h,d}\\
0& 0&                                  0&   0&                               0&      0&             0&   0
\end{bmatrix},
}
\end{equation}
where
\begin{align}
E_{0,2}=&-K_{t,p}+l_{t,1}mg+l_{t,2}(m_n+m_h)g\\
E_{3,4}=&-K_{n,p}+l_{n,1}(m_n+m_h)g+l_{n,2}m_h g\\
E_{6,6}=&- K_{h,p}+l_h m_h g
\end{align} 
with $m=m_t+m_n+m_h$

$L$ is an invertible matrix, which leads to the expression:
\begin{equation}
z=-L^{-1}E\xi_u-L^{-1}F\ddot{p} \label{eq:annexz}
\end{equation}

On another hand, we have that 
\begin{equation}
\dot{\xi_u}=Gz+M\xi_u+H\ddot{p}\label{eq:annexdotxiz}
\end{equation}
with 
\begin{equation}
G = \begin{bmatrix}
           0& 0& 0& 0& 0& 0& 0& 0\\
           0& 0& 0& 0& 0& 0& 0& 0\\
           0& 0& 0& 0& 0& 0& 0& 0\\
           1& 0& 0& 0& 0& 0& 0& 0\\
           0& 0& 0& 0& 0& 0& 0& 0\\
           0& 0& 1& 0& 0& 0& 0& 0\\
           0& 0& 0& 0& 0& 0& 0& 0\\
           0& 0& 0& 0& 0& 1& 0& 0
\end{bmatrix},  
\end{equation}
\begin{equation}       
H = \begin{bmatrix}
       1&
       0&
       0&
       0&
       0&
       0&
       0
\end{bmatrix}^\top,
\end{equation}
and
\begin{equation}      
M = \begin{bmatrix}    
           0& 1& 0& 0& 0& 0& 0& 0\\
           0& 0& 0& 0& 0& 0& 0& 0\\
           0& 0& 0& 1& 0& 0& 0& 0\\
           0& 0& 0& 0& 0& 0& 0& 0\\
           0& 0& 0& 0& 0& 1& 0& 0\\
           0& 0& 0& 0& 0& 0& 0& 0\\
           0& 0& 0& 0& 0& 0& 0& 1\\
           0& 0& 0& 0& 0& 0& 0& 0
\end{bmatrix}
\end{equation}

Replacing (\ref{eq:annexz}) in (\ref{eq:annexdotxiz}) provides us finally with the following linear dynamics:
\begin{align}
\dot{\xi}_u=(M-GL^{-1}E)\xi_u+(H-GL^{-1}F)\ddot{p},
\end{align}
which is rewritten as following :
\begin{equation}
\dot{\xi}_u=A\xi_u+B\ddot{p} 
\end{equation}

\subsection{Rigid upper-body}
For the case of rigid neck model, the state is much simpler since there is only one segment in the upper-body. In this case, $z = \begin{pmatrix} \ddot{\alpha} & f_t \end{pmatrix}^\top$ and $\xi_u=\begin{pmatrix}p & \dot{p} & \alpha & \dot{\alpha}\end{pmatrix}^\top$ Nevertheless, all the equations above are the same except that the constant matrices have different values:
\begin{equation}
L = \begin{bmatrix}     -I& -l_1\\
         mI&  -1\\
      \end{bmatrix}
\end{equation}
where $l_1=\frac{1}{m}\left(m_t\frac{l_t}{2}+m_n(l_t+\frac{l_n}{2})+m_h(l_t+l_n+l_h)\right)$ is the distance from the hip joint to the center of mass of the upper-body, $I=m_t(l_1-\frac{l_t}{2})^2 + m_n(l_1-l_t-\frac{l_n}{2})^2+ m_h(l_1-l_t-l_n-l_h)^2$ is the moment of inertia of the upper-body relative to its CoM.

\begin{equation}
E = \begin{bmatrix}     0& 0& -K_{t,p}+l_1m& -K_{t,d}\\
           0& 0&        0&   0
      \end{bmatrix}
\end{equation}

\begin{equation}
F = \begin{bmatrix} 0&
       m
      \end{bmatrix}^\top
\end{equation}

\begin{equation}
G = \begin{bmatrix} 0& 0\\
       0& 0\\
       0& 0\\
       1& 0
        \end{bmatrix}
\end{equation}

\begin{equation}
H = \begin{bmatrix} 0&
       1&
       0&
       0
      \end{bmatrix}^\top
\end{equation}

\begin{equation}
M = \begin{bmatrix}     0& 1& 0& 0\\
           0& 0& 0& 0\\
           0& 0& 0& 1\\
           0& 0& 0& 0
      \end{bmatrix}
\end{equation}   

\subsection{Remark}
It is worth to note that with these developments, it is easy to obtain the internal forces $f_i$ since they are components of the $z$ variable. The external force applied by the cart at the hip joint is simply the joint force $f_t$. This enables us to compute easily the force and energy required to move the cart which provides us with the results showed in the main text.

\section{The study of the sensitivity to random control parameters}
\subsection{The generation of random control parameters}
In order to study the sensitivity of the dynamical effects of head stabilization to the control parameters of these walking systems, we generate random values for these parameters. The considered parameters are the stiffness and damping of the support foot $K_{toe,p}$, $K_{toe,d}$, the impulsion velocity reference $\dot{l}_{p,r}$, the proportional and derivative gains for the interleg angle $K_{hip,p}$, $K_{hip,d}$ and for the trunk orientation $K_{t,p}$ and $K_{t,d}$. Each of these parameters was ranging between the quarter to the double or the triple of its initial value. The initial values are the values used to produce the results for the MFPTs in Figure~\ref{PlotIdeal}. Please see Table~\ref{tbl:randomvalues} for detailed values of these parameters.

\begin{table}
\begin{center}
\caption{
The range of values for sampling random parameters.}
\begin{tabular}{|c|c|c|c|}
\hline
&initial & min &  max\\
\hline 
$K_{toe,p}$ & 50000~N/m  &12500~N/m & 150000~N/m\\
$K_{toe,d}$ & 2000~N  & 500~N & 6000~N\\
$\dot{l}_{p,r}$ & 1~m/s& 0.25~m/s & 3~2m/s \\
$K_{hip,p}$ &  10~Nm/rad & 2.5~N m/rad & 30~N m/rad\\
$K_{hip,d}$  &1.5~Nms/rad &0.375~Nms/rad &4.5~Nms/rad\\ 
$K_{t,p}$  &300~Nm/rad &75~Nm/rad &600~Nm/rad\\ 
$K_{t,d}$  &150~Nms/rad &37.5~Nms/rad &300~Nms/rad\\ 
\hline 
\end{tabular}
\label{tbl:randomvalues}
\end{center}
\end{table} 

\subsection{The evaluation}
There are two parameters that were estimated for this study. The first one was the viability and the second one was the robustness to external disturbances.

The viability was estimated by starting the simulation on a flat floor at the limit cycle of the initial values, and stopping either at a fall or after 100 steps. The walker is considered viable if it does not fall.

The robustness was estimated by continuing the simulation after this 100th step. Then the ground inclination started to vary with 0.3 rad of standard deviation. The MFPT of this walker the average number of steps the walker is able to perform after the ground inclination started changing.

\end{appendix}

\begin{footnotesize}
\bibliographystyle{abbrvnat}

\bibliography{biblio}
\end{footnotesize}

\end{document}